\documentclass[11pt]{article}

\usepackage{acl}

\usepackage{xcolor}
\usepackage{rotating}
\usepackage{multirow}
\usepackage{xspace}
\usepackage{latexsym}
\usepackage{graphicx}
\usepackage{hyperref}
\usepackage{array}
\usepackage{color, colortbl}
\usepackage[inline]{enumitem}
\usepackage[flushleft]{threeparttable}

\usepackage{authblk}
\usepackage{times}
\usepackage[T1]{fontenc}

\usepackage[utf8]{inputenc}

\usepackage{microtype}

\long\def\eat#1{}

\def\dataset{\textsc{XtremeSpeech}\xspace}

\def\lime{{\fontfamily{lmr}\selectfont LIME}\xspace}
\def\extremity{{\fontfamily{lmr}\selectfont EXTREMITY}\xspace}
\def\target{{\fontfamily{lmr}\selectfont TARGET}\xspace}
\def\removal{{\fontfamily{lmr}\selectfont REMOVAL}\xspace}

\title{Listening to Affected Communities to Define Extreme Speech:\\
	Dataset and Experiments}

\author{\textbf{Antonis Maronikolakis}$^{1^*}$ ~ \textbf{Axel Wisiorek}$^{1,2}$ ~ \textbf{Leah Nann}$^3$ ~ \textbf{Haris Jabbar}$^1$ \\\textbf{Sahana Udupa}$^3$ ~ \textbf{Hinrich Schütze}}

\affil[ ]{$^1$CIS, Center for Information and Language Processing }
\affil[ ]{$^2$Center for Digital Humanities ~ $^3$Institute of Social and Cultural Anthropology}
\affil[ ]{LMU Munich}
\affil[ ]{\textit{$^*$antmarakis@cis.lmu.de}}

\def\figref#1{Figure~\ref{fig:#1}}
\def\figlabel#1{\label{fig:#1}\label{p:#1}}

\def\tabref#1{Table~\ref{tab:#1}}

\def\tablabel#1{\label{tab:#1}\label{p:#1}}

\def\secref#1{\S\ref{sec:#1}}
\def\seclabel#1{\label{sec:#1}}

\newcounter{notecounter}
\newcommand{\enotesoff}{\long\gdef\enote##1##2{}}
\newcommand{\enoteson}{\long\gdef\enote##1##2{{
		\stepcounter{notecounter}
		{\large\textbf{ \hspace{1cm}\arabic{notecounter} $<<<$ ##1: ##2 $>>>$\hspace{1cm}}}}}}
\enoteson
\enotesoff                      

\definecolor{myblue}{rgb}{0,0,.5}
\definecolor{myred}{rgb}{1,0,0}
\definecolor{mypurple}{rgb}{.5,0,.5}

\begin{document}
\maketitle
\begin{abstract}
	Building on current work on multilingual hate speech (e.g., \citet{multi_hate_speech}) and 
	hate speech reduction (e.g., \citet{social_bias_frames}), we
	present \dataset,\footnote{Code and data available at \url{https://github.com/antmarakis/xtremespeech}} a new hate speech dataset containing 20,297 social
	media passages from Brazil, Germany, India and Kenya. The
	key novelty is that we directly involve the affected communities in collecting and
	annotating the data --
	as
	opposed to giving companies and governments control over
	defining and combatting hate speech.  This inclusive approach results in
	datasets more representative of actually occurring
	online speech and is likely to facilitate the
	removal of the social media content that marginalized communities
	view as causing the most harm.  Based on \dataset,
	we establish novel tasks with accompanying
	baselines, provide evidence that cross-country training is
	generally	not feasible due to cultural differences between countries
	and perform an interpretability analysis of BERT's
	predictions.
\end{abstract}

\enote{hs}{uncomment
  setmainfont{NimbusRomNo9L-Regu}
}

\section{Introduction}
\seclabel{intro}
Much effort has been devoted to curating data in the area of
hate speech, from foundational work \cite{waseem,davidson}
to more recent, broader \cite{social_bias_frames} as well as
multilingual \cite{multi_hate_speech} approaches. However,
the demographics of those targeted by hate speech and those creating datasets are often quite different.
For example, in \citet{founta}, 66\% of
annotators are male and in \citet{social_bias_frames}, 82\%
are white.
This may lead to unwanted bias (e.g.,
disproportionately labeling African American English as hateful
\cite{aae_bias_hatespeech,racial_bias_in_data}) and to
collection of data
that is not representative of the comments directed at
target groups; e.g., a white person may not see and not have access to
hate speech targeting a particular racial group.

An example from our dataset is the Kenyan social media post
``\ldots\ We were taught that such horrible things can only be found
in Luo Nyanza.''  The Luo are an ethnic group in Kenya;
Nyanza is a Kenyan province. The post is incendiary because
it suggests that the Luo are responsible for horrible
things, insinuating that retaliation against them may be
justified. Only a group of people deeply rooted in Kenya can
collect such examples and understand their significance.

\begin{figure}[!t]
	\centering
	\includegraphics[scale=0.15]{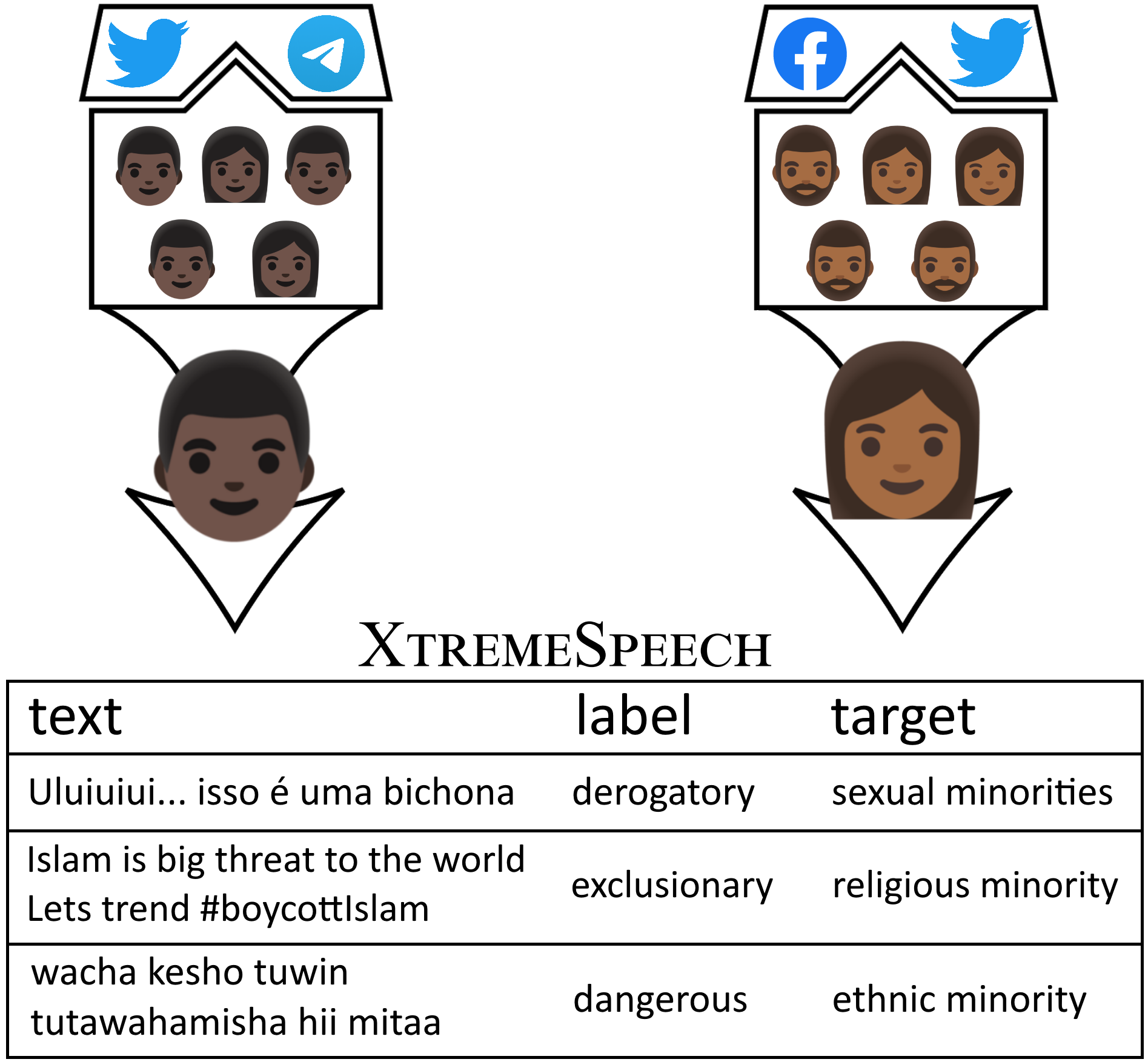}
	\caption{Overview of hate speech data
		collection. Instead of querying for data
		on our own, we work with fact-checkers
		advocating
		for targeted communities who collect and label
		data as they organically come across it.
		This inclusive approach
		results in datasets more representative of
		online speech the
communities are exposed to.
See \secref{extremedef} for definition of \dataset labels.}
	\figlabel{pitch}
\end{figure}

\textbf{\dataset.}
In this paper,
we
present \dataset, a new hate speech dataset containing 20,297 social
media passages from Brazil, Germany, India and Kenya. The
key novelty is that we empower the local affected communities (as
opposed to companies and governments) to collect and
annotate the data, thus avoiding the problems inherent in
approaches that hire outside groups for hate speech dataset
creation.
In more detail, we built a team of annotators from
fact-checking groups from the four different
countries. These annotators both collected and annotated
data from channels most appropriate for their respective
communities. They were also involved in all phases of the
creation of \dataset, from designing the annotation scheme to labeling.
Our inclusive approach results in
a dataset
that  better represents
content targeting these communities and that minimizes
bias against them because fact-checkers are trained to be
objective and know the local context.
\figref{pitch} gives a high-level overview of data
collection and annotation for \dataset.

\dataset also is a valuable resource because 
existing hate speech
resources are not representative for problematic speech on a
worldwide scale:
they mainly cover Western democracies. In contrast, our
selection is more balanced, containing three countries from
the Global South and one Western democracy.

We present  a data statement
(see \citet{data_statements})
in Appendix \ref{app:data_statement}.

\textbf{Anthropological perspective.}  It has been argued
that the NLP community does not sufficiently engage in
interdisciplinary work with other fields that address
important aspects of hate speech
\cite{collecting_sociocultural_data}.
In this work, we take an anthropological
perspective: the research we present is a collaboration of
anthropologists and computational
linguists.
\eat{As the discipline that studies people's
customs
and institutions
in the context of their society and culture,
sociocultural anthropology}
As a discipline that engages in the study of society and culture by exploring the lived worlds of people, and with a commitment to the application of knowledge to address human problems, sociocultural anthropology
can provide a highlevel framework
for investigating and theorizing about the phenomenon of
hate speech and its cultural variations.


We also take an anthropological perspective for
defining the terminology in this paper.
Potentially harmful online speech
is most often referred to
by NLP researchers and
general
media\footnote{\href{https://items.ssrc.org/disinformation-democracy-and-conflict-prevention/classifying-and-identifying-the-intensity-of-hate-speech/}{https://items.ssrc.org/disinformation-democracy-and-conflict-prevention/classifying-and-identifying-the-intensity-of-hate-speech/}} as \textbf{hate
	speech}.
From its original, culturally-grounded meaning, \emph{hate speech}
has evolved into a primarily
legal and
political term
with different definitions, depending on who uses it
\cite{hate_speech_high_court_decisions,white_paper,cyberhate_laws}.
We therefore
use the concept of
\textbf{extreme speech} from anthropology
and adopt its definition
as \emph{speech that pushes the boundaries of
	civil language} \cite{extreme_speech_sahana,policy_brief}.
In investigating extreme speech,
anthropologists
focus on cultural variation and historical conditions that
shape harmful speech.


\textbf{Extreme speech categories.}
We differentiate between extreme
speech
that 
requires removal (\textbf{denoted
	\textsf{R}}) and speech for which
moderation  (\textbf{denoted
	\textsf{M}}) is sufficient.
Extreme speech of the \textsf{M} category consists of
\textbf{derogatory} speech -- roughly, disrespectful and
negative comments about a group that are unlikely to directly
translate into specific harm. We further subdivide \textsf{R} extreme speech into
\textbf{exclusionary} extreme speech (roughly: speech 
inciting discrimination) and
\textbf{dangerous} extreme speech (roughly: speech inciting
violence); definitions are given in \secref{extremedef}.
This distinction is important when considering removal of extreme speech; e.g.,
dangerous speech may warrant more immediate and drastic
action than  exclusionary speech.

\dataset does not contain neutral
text, focusing solely on 
\textsf{M} and \textsf{R}
extreme speech. Neutral text has
been shown to be easier to label both for humans and models
while identifying and subclassifying
non-neutral text (i.e.,
extreme speech) remains the Achilles'
heel of NLP models \cite{davidson,hatespeech_crosslingual_embds}.

Finally, we also annotate the targets of extreme speech;
examples
are ``religious minorities'' and ``immigrants'' (frequent
targets in India and Germany, respectively).

\textbf{Classification tasks.}
We define three classification tasks. (i) \textbf{\removal.}
The two-way 
classification \textsf{M} vs.\ \textsf{R}.
(ii) \textbf{\extremity.}
The three-way 
classification according to degree of extremity: derogatory vs.\ exclusionary vs.\ dangerous.
(iii) \textbf{\target.}
Target group classification.

We propose a series
of baselines and 
show that model performance
is mediocre for \removal, 
poor for \extremity and good for \target.
Further, we show
that \texttt{BERT}-based models are unable to generalize in
cross-country and cross-lingual settings, confirming
the intuition that cultural and world knowledge is needed for
this task. We also perform a model interpretability
analysis with LIME \cite{lime} to uncover potential model
biases and deficiencies.


\textbf{Contributions.} In summary, we  \begin{enumerate*}[label={\textbf{(\roman{*})}}]
	\item establish a community-first framework of data curation,
	\item present \dataset, a dataset of 20,297 \textit{extreme
		speech} passages from Brazil, Germany, India and
	Kenya, capturing target groups and multiple levels of extremity,
	\item propose a series of tasks and baselines,
	as the basis for meaningful comparison with future work,
	\item show performance both for models and humans is low across tasks except in target group classification,
	\item confirm the intuition that extreme speech is dependent on social and cultural knowledge, with low cross-lingual and cross-country performance
\end{enumerate*}.

\section{Related Work}

Earlier work in hate speech detection focused on data
collection,
curation and annotation
frameworks \cite{waseem,davidson,founta}. Recent work has
expanded the set of captured labels to include more
pertinent information such as targets and other forms of
abuse
\cite{social_bias_frames,incivility_in_news,misogyny_annotated_data,hate_towards_political_opponent,refugee_crisis_german}
as well as benchmarking
\cite{hatecheck,hatexplain}. Analysis of datasets has been
performed too
\cite{hate_speech_data_analysis,intersectional_bias,hatespeech_biased_datasets,hate_speech_cross_dataset,racial_bias_in_data}.

Work has also been conducted to expand research to multiple
languages
\cite{multi_hate_speech,hatespeech_crosslingual_embds,refugee_crisis_german,hatespeech_zero_shot_cross_lingual,zoph-etal-2016-transfer,african_low_resource,nekoto-etal-2020-participatory}. \dataset
contributes to this goal by providing Brazilian Portuguese, German, Hindi and Swahili data.

Research has also been conducted to investigate annotation bias
and annotator pools
\cite{annotator_bias_demographic,are_you_racist,refugee_crisis_german,ethical_implications_crowdsourcing,global_workforce_demographics},
as well as bias (especially racial) in existing
datasets 
\cite{racial_bias_in_hatespeech,civil_text_rephrasing}. It
was found that data can reflect and propagate annotator
bias. To address this, we diversify the
annotator pool in our work.

In another line of work, theoretical foundations are being established, in the form of taxonomies \cite{hatespeech_taxonomy}, definitions \cite{implicitly_abusive_language,understanding_abuse} and theory \cite{unhealthy_convo_attributes,datafication_of_hate}. We are adding to this  with definitions based on fieldwork and grounded research, inspired by anthropological and ethnographic work that investigates the societal impact of online hate and extreme speech \cite{boromisza,stop_the_presses,bolivia,extreme_speech_sahana,denmark}.

Further, strides have been made in the ethics of AI. Who should collect data and who is responsible for model deployment decisions? Calls have been made for more inclusive pools of annotators and domain experts overseeing NLP projects, as well as exploration of other ethical dilemmas (\citet{convenience_or_death,collecting_sociocultural_data,diversity_n_inclusion_metrics,hate_speech_frontiers,gebru_gender_race,decolonial_ai_theory}, \textit{inter alia}). With our focus on community-embedded fact-checkers our framework is more inclusive than previous work.

\section{Dataset}

\subsection{Dataset Description}
\seclabel{datasetdesc}

\dataset consists of 20,297 passages, each targeted at one or more groups (e.g., immigrants). Data is collected from Brazil, Germany, India and
Kenya. Passages are written in Brazilian Portuguese, German, Hindi and
Swahili, as well as in English.
English can either
be used on its own, or in conjunction with the local
language in the form of code switching. We capture this
in the annotation: passages that contain English
-- even if it is only a hashtag in a tweet -- are marked as containing both
languages. \tabref{lang_dist} shows the distribution of languages.

Further, \dataset is platform-agnostic, with text collected
from multiple online platforms, as well as direct messaging
(anonymized) from the third quarter of 2020 until the end of
2021. In more detail, Brazilian annotators sourced WhatsApp groups, the German team collected data from Facebook, Instagram, Telegram, Twitter and YouTube, Indian annotators sourced Facebook and Twitter and the Kenyan annotators collected data from Facebook, Twitter and WhatsApp. While forms of extreme speech may originate from one place, dissemination to other platforms is swift \cite{deplatforming}. Proprietary efforts have also taken a platform-agnostic approach.\footnote{\url{https://www.perspectiveapi.com/}}

Passages were labeled both on content and target levels. On
their content they are labeled as derogatory,
exclusionary or
dangerous. On the target level, we make a distinction between text targeted at protected groups and at institutions of power. We take into account the following protected groups: ethnic minorities, immigrants, religious minorities, sexual minorities, women, racialized groups, historically oppressed caste groups, indigenous groups and large ethnic groups. We also give the annotators the option to input any other group. For institutions of power, possible targets are politicians, legacy media and the state. To allow for political discourse, extreme speech against institutions of power should not be filtered out, so such speech was marked as derogatory.

\subsection{Extreme Speech Definitions}
\seclabel{extremedef}

Building on \citet{dangerous_speech} and \citet{counter_online_hate}, we define extreme speech labels as follows:\footnote{Definitions were shared as annotation instructions.}

\textbf{Derogatory Extreme Speech}: Text that crosses the boundaries of civility within specific contexts and targets either individuals/groups based on protected characteristics (e.g., ethnicity and religious affiliation) or institutions of power (state, media, politicians). Includes derogatory expressions about abstract categories/concepts.

\textbf{Exclusionary Extreme Speech}: Text that calls for or implies exclusion of vulnerable groups based on protected attributes (for example, ethnicity, religion and gender). Exclusionary text marginalizes, delegitimizes and discriminates against target groups. Text targeting abstract concepts or institutions is not exclusionary, except when there is reason to believe that such attacks call for or imply the exclusion of vulnerable groups associated with these abstract concepts or institutions.

\textbf{Dangerous Extreme Speech}: Text that has a reasonable chance to trigger harm against target groups (e.g., ostracism and deportation). All the following criteria should be met for a passage to be classified as dangerous: \begin{enumerate*}[label={(\roman{*})}]
	\item content calls for harm,
	\item speaker has high degree of influence over audience,
	\item audience has grievances and fears that the speaker can cultivate,
	\item target groups are historically disadvantaged and vulnerable to harm,
	\item influential means to disseminate speech
\end{enumerate*}.

Whereas derogatory extreme speech is a form of speech that
\textit{requires moderation but, generally, not removal} (denoted with
\textsf{M}), exclusionary and dangerous speech are forms of speech that do
\textit{require removal} (denoted with \textsf{R}) in most
cases to
protect users from potential harm. We make a distinction
between exclusionary and dangerous speech in order to
introduce a more fine-grained scale of extremity that can
dictate more focused policy (e.g.,
more severe punishment may be appropriate for dangerous speech). It has been shown in previous work that while
neutral text is easier to detect \cite{davidson,hatespeech_crosslingual_embds,risch-krestel-2020-bagging}, models find
it hard to differentiate between different types of extreme
speech (e.g., between our definitions of \textsf{M} or
\textsf{R}, or between merely offensive versus hateful speech), a task challenging even for humans.
By focusing on the difficult distinctions within non-neutral text,
we hope to contribute to research that will be able to
classify types of potentially harmful speech correctly in the future,
which is both the critical point
of extreme speech research and the main obstacle towards
effective filtering.

Exemplary cases for the three labels (derogatory,
exclusionary, dangerous)
were discussed in detail with the
annotators.
We believe our interdisciplinary approach
will lead to data more aligned with the real world and will
benefit the target groups and communities to greater effect.

\subsection{Data Collection}

\subsubsection{Annotator Profiles}

We selected  Brazil, Germany, India and Kenya to cover a range of cultures and communities. Each annotator is a fact-checker who i) is local,  ii) is
independent  (i.e., not employed by social media companies
or large media corporations) and iii) investigates the
veracity of news articles, including articles directed at or related to local communities. There are 8 female and 5 male annotators (per country, female/male counts are 2/1 in Brazil, 4/0 in Germany, 2/2 in India and 0/2 in Kenya).

Fact-checking companies were scouted and individual fact-checkers interviewed by our anthropology team to verify their familiarity with extreme speech, their expertise in local community affairs and their ability to act as annotators in our project.

We see independent fact-checkers  as a key stakeholder community that provides a feasible and meaningful gateway into cultural variation in online extreme speech. Through their job as fact-checkers, they regularly come in contact with extreme speech, with communities that peddle extreme speech as well as with communities targeted by extreme speech (further details in Appendix \ref{app:annotation}).

\subsubsection{Annotation Scheme}

Through an online interface, data is entered as found in
online media. This
interface (in the form of a web page, see
Appendix \ref{app:online_interface}) serves both as the data
entry point and the annotation form. After finding a passage
of extreme speech, annotators enter it in our form
and are prompted to label it (see categories in \secref{datasetdesc}).

\eat{Input from the annotators is taken into
	account in all stages of this process, to ensure smooth and
	reliable data collection.}

\subsection{Inter-annotator Agreement}

To verify the quality of \dataset, we calculate
inter-annotator agreement. The
data collected from one annotator is shown to another for
verification (details in Appendix
\ref{app:cross_annotation}).
Only the text passage is shown to annotators, without
prior category assignments.
The agreement scores we measure are:
Cohen's kappa ($\kappa$, \citet{kappa}),
Krippendorff's alpha ($\alpha$, \citet{alpha}), intraclass
correlation coefficient (two-way mixed, average score
$\mbox{ICC}(3, k)$ for $k=2$, \citet{icc}) and accuracy (defined as the percentage of passages where both annotators agreed).

For the three extreme speech labels, $\kappa = 0.23$,
$\alpha=0.24$ and $\mbox{ICC}(3,k)= 0.41$
(considered ``fair'' \cite{icc}). Accuracy is 63\% overall,
78\% for derogatory, 40\% for exclusionary and 19\% for
dangerous.  For the \textsf{M} vs.\ \textsf{R} task, accuracy
is 78.4\% for \textsf{M} and 46.3\% for \textsf{R}. For
the classification of the target of extreme speech, $\kappa = 0.69$.

Scores are  low compared to other NLP tasks,
which is unfortunately a widespread phenomenon in
hate speech research. In \citet{founta}, only in
55.9\% of passages did at least 4 out of 5 annotators
agree. In \citet{social_bias_frames}, the $\alpha$ score was
0.45, with a 76\% agreement on ``offensiveness'' and 74\% on ``targeted group''. In
\citet{davidson}, there was a 90\% agreement on whether text was neutral, offensive, or hateful. In
\citet{refugee_crisis_german}, a German dataset, $\alpha$
was between 0.18 and 0.29, while in
\citet{multi_hate_speech}, a multilingual dataset, $\alpha$
was between 0.15 and 0.24.

We argue that in our work, not only are we dealing with a
heavily imbalanced dataset, but also that the task is even
more challenging than prior work, which
collects both neutral passages and hate speech (e.g., in
\citet{davidson}). We only collect extreme speech, so whereas in prior work
the annotators need to differentiate between neutral and
extreme speech (a relatively easier task \cite{hatespeech_crosslingual_embds,risch-krestel-2020-bagging}), 
our annotators only make decisions on the hard task of
determining different degrees of extremity.

\subsection{Reannotation}
\seclabel{reannotation}

After discussing inconsistently labeled passages with
the annotators, we found that there was 
disagreement about groups currently in
power, specifically, the Kikuyu and Kalenjin ethnic groups
(more information in Appendix \ref{app:reannotation}). One
annotator considered them  ethnic minorities because most
other ethnic groups are pitted against them. The other
annotator did not view them as minorities because they
are (i) the two most populous ethnic groups and (ii) are not
in the minority when it comes to representation in positions of
power. A consensus was
reached to add a new target label, ``large ethnic group'',
to correctly represent this state of affairs in the
annotation.

As is common practice,
instead of limiting the reannotation to passages the
annotators disagreed on, we provided all potentially
affected passages for reannotation, i.e., all ``indigenous
group'' and ``ethnic minority'' passages.

\subsection{Dataset Analysis}

\subsubsection{Extreme Speech Analysis}
\dataset contains 20,297 passages from the four
countries. From each country, we chose to only collect data
on one local language plus English. The distribution of
languages is shown in \tabref{lang_dist}. While for
Germany and Brazil, English is rarely used, in India and
Kenya it is more prominent, both on its
own and in code switching.

\begin{table}[]
	\centering
	\small
	\begin{tabular}{lrrrr|}
		\cline{2-5}
		& \multicolumn{1}{|c}{\cellcolor[HTML]{9B9B9B}Brazil} & \cellcolor[HTML]{9B9B9B}Germany & \cellcolor[HTML]{9B9B9B}India & \cellcolor[HTML]{9B9B9B}Kenya \\
		\multicolumn{1}{|l}{\cellcolor[HTML]{C0C0C0}Local} & 5109 & 4922 & 2778 & 405 \\
		\multicolumn{1}{|l}{\cellcolor[HTML]{C0C0C0}English} & 0 & 6 & 1056 & 2695 \\
		\multicolumn{1}{|l}{\cellcolor[HTML]{C0C0C0}Both} & 0 & 71 & 1174 & 2081 \\ \hline
	\end{tabular}
	\caption{\dataset passages per country and language combination}
	\tablabel{lang_dist}
\end{table}

The distribution of labels,  shown in
Table \ref{labels_dist}, 
varies a lot from country to country. For example, in Germany annotators labeled far fewer passages as dangerous speech,
reflecting stricter regulatory controls over speech compared
to the other countries. Data is also heavily
imbalanced in Brazil, with the majority of passages being
derogatory.

The distribution of targets per country (shown in Table
\ref{target_dist}) again shows large divergences between
countries.
\eat{As a category that was added posthoc (see
	\secref{reannotation}),
	``large ethnic groups'' only occurs for Kenya.}
In Germany, immigrants are the main target
group because of right-wing opposition to 
recent immigration. In India,
religious minorities dominate the target group statistics
because of
the conflict between
Hindus and  Muslims.
Thus \dataset reflects a country's social and political
situation to a reasonable extent.

\begin{table}[]
	\centering
	\small
	\begin{tabular}{lrrrrr|}
		\cline{2-6}
		\multicolumn{1}{c|}{} & \cellcolor[HTML]{9B9B9B}Brazil & \cellcolor[HTML]{9B9B9B}Germany & \cellcolor[HTML]{9B9B9B}India & \cellcolor[HTML]{9B9B9B}Kenya & \cellcolor[HTML]{9B9B9B}\textit{Total} \\
		\multicolumn{1}{|l}{\cellcolor[HTML]{C0C0C0}Der.} & 4774 & 2643 & 2225 & 3389 & \textit{13031} \\
		\multicolumn{1}{|l}{\cellcolor[HTML]{C0C0C0}Exc.} & 115 & 2340 & 1422 & 1024 & \textit{4901} \\
		\multicolumn{1}{|l}{\cellcolor[HTML]{C0C0C0}Dan.} & 220 & 16 & 1361 & 768 & \textit{2365} \\ \hline
	\end{tabular}
	\caption{Distribution of extreme speech labels in \dataset (Der = Derogatory, Exc = Exclusionary, Dan = Dangerous)}
	\label{labels_dist}
\end{table}

\subsubsection{Word Frequency}
\label{sec:word_freq}

\tabref{words_freq} shows the
most frequent words for the three extreme
speech labels for the four countries.
We see 
that words indicative of sociopolitical conflict
appear frequently:
``comunista'' and ``feminista'' in Brazil;
``merkel'' (a German politician) and
``deutsche'' (meaning: ``German''), as well as the word for Jew,
``jude'' in Germany; words
referring to religion
(e.g.,
``muslims'', ``hindus'') in India. In
Kenya, political entities (``Ruto'' and ``Raila'', names of two Kenyan politicians) as well as
ethnic groups (e.g., ``Kikuyus'', ``Kalenjins'', two powerful groups in Kenya)
are among the most frequent words,
with
ethnic groups appearing particularly prominently in the two forms of extreme
speech that should be removed (\textsf{R}).

\begin{table*}[]
	\centering
	\scriptsize
	\begin{tabular}{ lp{3.1cm}p{3.1cm}p{3.1cm}p{3.1cm}|  }
		\cline{2-5}
		& \multicolumn{1}{|c}{\cellcolor[HTML]{9B9B9B}Brazil} & \multicolumn{1}{c}{\cellcolor[HTML]{9B9B9B}Germany} & \multicolumn{1}{c}{\cellcolor[HTML]{9B9B9B}India} & \multicolumn{1}{c|}{\cellcolor[HTML]{9B9B9B}Kenya} \\
		\multicolumn{1}{|l}{\cellcolor[HTML]{C0C0C0}Der.} & puta, vai, filho, arrombada, pra, vc, comunista, cu, traveco, tomar & mehr, deutschland, merkel, schon, mal, ja, immer, deutsche, land, neger & \textit{ke}, \textit{nahii}, muslims, \textit{bhiimaTa}, muslim, \textit{mulla}, hindu, india, \textit{desha}, hindus & Ruto, people, Raila, know, ruto, Kenya, never, even, Uhuru, us \\ [0.625cm]
		\multicolumn{1}{|l}{\cellcolor[HTML]{C0C0C0}Exc.} & puta, feminista, pra, bichona, ucranizar, nojenta, ser, marmita, bandido, cu & deutschland, mehr, darf, ja, antwort, land, deutschen, juden, deutsche, mal & muslims, hindu, \textit{desha},  bhimte, india,  \textit{bhiima}, hindus, \textit{bhaarata}, \textit{mulla}, country & Kikuyus, Ruto, Kenya, kikuyu, Raila, people, never, Uhuru, Luos, Kalenjins \\ [0.65cm]
		\multicolumn{1}{|l}{\cellcolor[HTML]{C0C0C0}Dan.} & fechar, stf, pra, povo, ucranizar, vai, q, ser, hora, bolsonaro & jude, europa, darf, juden, muslim, scheiss, freiheitskampf, völker, fällt, niemals & muslims, muslim, hindu, hindus, india, girls, love, \textit{desha}, women, religion & Ruto, people, killed, Kikuyus, Raila, Kenya, know, Rift, must, time \\ \hline
	\end{tabular}
	\caption{Most frequent words per label and country in \dataset. Hindi text originally in the Devanagari script was converted to Latin characters (in italics) to comply with arxiv's typesetting policy.}
	\tablabel{words_freq}
\end{table*}

\section{Experiments}
\seclabel{experiments}

We establish \dataset baselines for large pretrained models
and traditional machine learning models (details in Appendix
\ref{app:models}).
As introduced in \secref{intro},
we address three novel tasks: predicting the extremity of speech (\extremity), whether a passage should be removed or not (\removal) and the target of extreme speech (\target).

Unless noted otherwise, our measure is micro-averaged F1.
We split each country set 80:10:10 into train:dev:test,
sampling equally for all labels.\footnote{The German subset
  only contains 16 dangerous passages, so results for
  dangerous speech are of limited utility.}
In Tables \ref{extreme_results}, \ref{acceptable_results}, \ref{target_classification}, \ref{cross_lingual}, \ref{cross_lingual_english} we show results on the development set (test set results in Appendix \ref{app:test_set}).

We evaluate both multilingual (\texttt{mBERT},
\texttt{XLM-R} \cite{xlmr}) and monolingual (\texttt{langBERT})
models. Each monolingual model was pretrained on the local
language we are using for each corresponding country; e.g., the
Indian model was pretrained on Hindi. For finetuning and classification with \texttt{BERT}-based models, a task-specific head is added that takes as input the [CLS] token representation.

\subsection{\extremity Task}
\seclabel{extremity}

Table \ref{extreme_results} shows that
baseline performance is rather low in three-way
classification (\extremity). In India and Kenya, performance is
acceptable; in Germany as well if we exclude the dangerous
label, which 
only has 16 passages. In Brazil, however, where the predominant class is derogatory speech (with more than 90\% of all passages labeled as derogatory), performance is low, with no model managing to detect exclusionary speech.

\texttt{XLM-R} performs relatively poorly, only scoring
competitively in the low-resource Kenyan
set. \texttt{langBERT} is competitive for Brazil and
Germany, less so for Kenya and performs badly for India.
This can be explained by the divergence of
pretraining and \dataset text:
all \texttt{langBERT}
models are pretrained on a single language (Brazilian Portuguese, German, Hindi and Swahili, respectively). In the Brazilian
and German sets there is primarily only one language used so
\texttt{langBERT} performs better in those sets, while it
performs worse in countries where English is more
predominant both as a standalone language and in code switching, which is the case for India and Kenya.

\eat{A major issue plaguing this area of research is that substantial world knowledge is required, especially around sociopolitical power dynamics, something that models struggle to capture \cite{learning_social_meaning,modeling_social_factors}.}

\subsection{\removal Task}

Table \ref{acceptable_results} shows that results are overall
better for the binary task \textsf{M} (moderation)
vs.\ \textsf{R} (removal) than for the fine-grained \extremity task. \texttt{BERT}-based models perform particularly well. \texttt{mBERT} performs especially well for India and the monolingual \texttt{langBERT} models again perform well for Brazil and Germany; this time we see improvements for Kenya too. LSTMs perform well, in some instances
competitively with transformers. \texttt{XLM-R} does not
seem to compute good representations and performs poorly
for all languages except for the low-resource Kenyan
dataset.

\subsection{\target Task}

Table
\ref{target_classification} shows that transformers are effective for 
the 8-way multilabel classification of target.
In \tabref{words_freq} and \tabref{lime_explainability}, we show top words according to
frequency in the dataset and contribution to \texttt{mBERT} predictions in the \extremity task, respectively.
Words denoting ethnicity (``kikuyu''), religion
(``hindu'', ``Muslim'') and gender (``puta'', ``girls'') appear
often and, not surprisingly, are reliable indicators of
targeted groups, making this task easier than the other two.

\begin{table*}[]
	\centering
	\small
	\begin{tabular}{lcccccccccc}
		\cline{2-6}
		\multicolumn{1}{c|}{} & \multicolumn{2}{c}{\cellcolor[HTML]{9B9B9B}Brazil} & \multicolumn{2}{c}{\cellcolor[HTML]{9B9B9B}Germany} & \multicolumn{2}{c}{\cellcolor[HTML]{9B9B9B}India} & \multicolumn{2}{c}{\cellcolor[HTML]{9B9B9B}Kenya} & \multicolumn{2}{c|}{\cellcolor[HTML]{9B9B9B}\textit{Total}} \\ \cline{2-9}
		\multicolumn{1}{l|}{\cellcolor[HTML]{FFFFFF}} & \multicolumn{1}{c}{\cellcolor[HTML]{EFEFEF}n} & \multicolumn{1}{c}{\cellcolor[HTML]{EFEFEF}\%} & \multicolumn{1}{c}{\cellcolor[HTML]{EFEFEF}n} & \multicolumn{1}{c}{\cellcolor[HTML]{EFEFEF}\%} & \multicolumn{1}{c}{\cellcolor[HTML]{EFEFEF}n} & \multicolumn{1}{c}{\cellcolor[HTML]{EFEFEF}\%} & \multicolumn{1}{c}{\cellcolor[HTML]{EFEFEF}n} & \multicolumn{1}{c}{\cellcolor[HTML]{EFEFEF}\%} & \multicolumn{1}{c}{\cellcolor[HTML]{EFEFEF}\textit{n}} & \multicolumn{1}{c|}{\cellcolor[HTML]{EFEFEF}\textit{\%}} \\
		\multicolumn{1}{|l}{\cellcolor[HTML]{C0C0C0}Religious Minorities} & \multicolumn{1}{r}{16} & \multicolumn{1}{r}{0.5} & \multicolumn{1}{r}{1269} & \multicolumn{1}{r}{23.8} & \multicolumn{1}{r}{3522} & \multicolumn{1}{r}{64.7} & \multicolumn{1}{r}{111} & \multicolumn{1}{r}{2.2} & \multicolumn{1}{r}{\textit{4918}} & \multicolumn{1}{r|}{\textit{25.4}} \\
		\multicolumn{1}{|l}{\cellcolor[HTML]{C0C0C0}Any Other} & \multicolumn{1}{r}{1066} & \multicolumn{1}{r}{30.5} & \multicolumn{1}{r}{34} & \multicolumn{1}{r}{0.6} & \multicolumn{1}{r}{356} & \multicolumn{1}{r}{6.5} & \multicolumn{1}{r}{1534} & \multicolumn{1}{r}{30.3} & \multicolumn{1}{r}{\textit{2990}} & \multicolumn{1}{r|}{\textit{15.5}} \\
		\multicolumn{1}{|l}{\cellcolor[HTML]{C0C0C0}Immigrants} & \multicolumn{1}{r}{28}& \multicolumn{1}{r}{0.8} & \multicolumn{1}{r}{2355} & \multicolumn{1}{r}{44.1} & \multicolumn{1}{r}{109} & \multicolumn{1}{r}{2.0} & \multicolumn{1}{r}{292} & \multicolumn{1}{r}{5.8} & \multicolumn{1}{r}{\textit{2784}} & \multicolumn{1}{r|}{\textit{14.3}} \\
		\multicolumn{1}{|l}{\cellcolor[HTML]{C0C0C0}Women} & \multicolumn{1}{r}{1479} & \multicolumn{1}{r}{42.3} & \multicolumn{1}{r}{367} & \multicolumn{1}{r}{6.9} & \multicolumn{1}{r}{418} & \multicolumn{1}{r}{7.7} & \multicolumn{1}{r}{396} & \multicolumn{1}{r}{7.8} & \multicolumn{1}{r}{\textit{2660}} & \multicolumn{1}{r|}{\textit{13.8}} \\
		\multicolumn{1}{|l}{\cellcolor[HTML]{C0C0C0}Large Ethnic Groups} & \multicolumn{1}{r}{0} & \multicolumn{1}{r}{0.0} & \multicolumn{1}{r}{0} & \multicolumn{1}{r}{0.0} & \multicolumn{1}{r}{0} & \multicolumn{1}{r}{0.0} & \multicolumn{1}{r}{2273} & \multicolumn{1}{r}{44.8} & \multicolumn{1}{r}{\textit{2273}} & \multicolumn{1}{r|}{\textit{11.8}} \\
		\multicolumn{1}{|l}{\cellcolor[HTML]{C0C0C0}Sexual Minorities} & \multicolumn{1}{r}{674} & \multicolumn{1}{r}{19.3} & \multicolumn{1}{r}{347} & \multicolumn{1}{r}{6.5} & \multicolumn{1}{r}{89} & \multicolumn{1}{r}{1.6} & \multicolumn{1}{r}{80} & \multicolumn{1}{r}{1.6} & \multicolumn{1}{r}{\textit{1190}} & \multicolumn{1}{r|}{\textit{6.2}} \\
		\multicolumn{1}{|l}{\cellcolor[HTML]{C0C0C0}\scriptsize{Historically Oppressed Caste Groups}} & \multicolumn{1}{r}{45} & \multicolumn{1}{r}{1.3} & \multicolumn{1}{r}{1} & \multicolumn{1}{r}{0.0} & \multicolumn{1}{r}{853} & \multicolumn{1}{r}{15.7} & \multicolumn{1}{r}{33} & \multicolumn{1}{r}{0.7} & \multicolumn{1}{r}{\textit{932}} & \multicolumn{1}{r|}{\textit{4.8}} \\
		\multicolumn{1}{|l}{\cellcolor[HTML]{C0C0C0}Racialized Groups} & \multicolumn{1}{r}{78} & \multicolumn{1}{r}{2.2} & \multicolumn{1}{r}{527} & \multicolumn{1}{r}{9.8} & \multicolumn{1}{r}{3} & \multicolumn{1}{r}{0.1} & \multicolumn{1}{r}{80} & \multicolumn{1}{r}{1.6} & \multicolumn{1}{r}{\textit{688}} & \multicolumn{1}{r|}{\textit{3.6}} \\
		\multicolumn{1}{|l}{\cellcolor[HTML]{C0C0C0}Ethnic Minorities} & \multicolumn{1}{r}{58} & \multicolumn{1}{r}{1.7} & \multicolumn{1}{r}{430} & \multicolumn{1}{r}{8.1} & \multicolumn{1}{r}{89} & \multicolumn{1}{r}{1.6} & \multicolumn{1}{r}{77} & \multicolumn{1}{r}{1.5} & \multicolumn{1}{r}{\textit{654}} & \multicolumn{1}{r|}{\textit{3.4}} \\
		\multicolumn{1}{|l}{\cellcolor[HTML]{C0C0C0}Indigenous Groups} & \multicolumn{1}{r}{50} & \multicolumn{1}{r}{1.4} & \multicolumn{1}{r}{6} & \multicolumn{1}{r}{0.1} & \multicolumn{1}{r}{5} & \multicolumn{1}{r}{0.1} & \multicolumn{1}{r}{195} & \multicolumn{1}{r}{3.8} & \multicolumn{1}{r}{\textit{256}} & \multicolumn{1}{r|}{\textit{1.3}} \\ \hline
	\end{tabular}
	\caption{Total number (n) and percentage (\%) of messages
	  directed at target groups in
        \dataset}
	\label{target_dist}
\end{table*}

\begin{table*}[]
	\centering
	\small
	\begin{tabular}{lrrrrrcrrrrrr|}
		\cline{2-13}
		\cellcolor[HTML]{FFFFFF} & \multicolumn{3}{|c}{\cellcolor[HTML]{9B9B9B}Brazil} & \multicolumn{3}{c}{\cellcolor[HTML]{9B9B9B}Germany} & \multicolumn{3}{c}{\cellcolor[HTML]{9B9B9B}India} & \multicolumn{3}{c|}{\cellcolor[HTML]{9B9B9B}Kenya} \\ \cline{2-13} 
		\multicolumn{1}{l|}{} & \multicolumn{1}{c}{\cellcolor[HTML]{EFEFEF}Der.} & \multicolumn{1}{c}{\cellcolor[HTML]{EFEFEF}Exc.} & \multicolumn{1}{c}{\cellcolor[HTML]{EFEFEF}Dan.} & \multicolumn{1}{c}{\cellcolor[HTML]{EFEFEF}Der.} & \multicolumn{1}{c}{\cellcolor[HTML]{EFEFEF}Exc.} & \multicolumn{1}{c}{\cellcolor[HTML]{EFEFEF}Dan.} & \multicolumn{1}{c}{\cellcolor[HTML]{EFEFEF}Der.} & \multicolumn{1}{c}{\cellcolor[HTML]{EFEFEF}Exc.} & \multicolumn{1}{c}{\cellcolor[HTML]{EFEFEF}Dan.} & \multicolumn{1}{c}{\cellcolor[HTML]{EFEFEF}Der.} & \multicolumn{1}{c}{\cellcolor[HTML]{EFEFEF}Exc.} & \multicolumn{1}{c|}{\cellcolor[HTML]{EFEFEF}Dan.} \\ \multicolumn{1}{|l}{\cellcolor[HTML]{C0C0C0}Human} & 97.2 & 21.2 & 0.0 & 73.0 & 61.6 & 0.0 & 91.1 & 16.9 & 4.9 & 68.9 & 10.7 & 57.2 \\ \multicolumn{1}{|l}{\cellcolor[HTML]{C0C0C0}Majority} & 100.0 & 0.0 & 0.0 & 100.0 & 0.0 & 0.0 & 100.0 & 0.0 & 0.0 & 100.0 & 0.0 & 0.0 \\ \multicolumn{1}{|l}{\cellcolor[HTML]{C0C0C0}SVM} & 100.0 & 0.0 & 35.6 & 67.8 & 62.9 & 0.0 & 76.7 & 29.8 & 65.6 & 89.6 & 41.9 & 38.8 \\
		\multicolumn{1}{|l}{\cellcolor[HTML]{C0C0C0}LSTM} & 98.4 & 0.8 & 0.0 & 59.4 & 68.6 & 0.0 & 56.3 & 64.8 & 0.0 & 64.9 & 63.4 & 0.0 \\
		\multicolumn{1}{|l}{\cellcolor[HTML]{C0C0C0}langBERT} & 99.7 & 0.0 & 54.8 & 62.0 & 70.6 & 0.0 & 87.4 & 0.0 & 53.4 & 83.3 & 38.5 & 45.2 \\
		\multicolumn{1}{|l}{\cellcolor[HTML]{C0C0C0}mBERT} & 98.9 & 0.0 & 49.3 & 56.3 & 72.4 & 0.0 & 60.9 & 45.5 & 81.3 & 83.5 & 48.4 & 48.8 \\
		\multicolumn{1}{|l}{\cellcolor[HTML]{C0C0C0}XLM-R} & 100.0 & 0.0 & 0.0 & 58.7 & 76.4 & 0.0 & 89.1 & 6.7 & 56.1 & 88.3 & 46.9 & 40.0 \\ \hline
	\end{tabular}
	\caption{F1 on dev for \extremity, the three-way extreme speech
		classification task}
	\label{extreme_results}
\end{table*}

\begin{table*}[]
	\centering
	\small
	\begin{tabular}{
			>{\columncolor[HTML]{C0C0C0}}crrrrrrrr|}
		\cline{2-9}
		\cellcolor[HTML]{FFFFFF} & \multicolumn{2}{|c}{\cellcolor[HTML]{9B9B9B}Brazil} & \multicolumn{2}{c}{\cellcolor[HTML]{9B9B9B}Germany} & \multicolumn{2}{c}{\cellcolor[HTML]{9B9B9B}India} & \multicolumn{2}{c|}{\cellcolor[HTML]{9B9B9B}Kenya} \\ \cline{2-9} 
		\multicolumn{1}{l|}{\cellcolor[HTML]{FFFFFF}} & \multicolumn{1}{c}{\cellcolor[HTML]{EFEFEF}\textsf{M}} & \multicolumn{1}{c}{\cellcolor[HTML]{EFEFEF}\textsf{R}} & \multicolumn{1}{c}{\cellcolor[HTML]{EFEFEF}\textsf{M}} & \multicolumn{1}{c}{\cellcolor[HTML]{EFEFEF}\textsf{R}} & \multicolumn{1}{c}{\cellcolor[HTML]{EFEFEF}\textsf{M}} & \multicolumn{1}{c}{\cellcolor[HTML]{EFEFEF}\textsf{R}} & \multicolumn{1}{c}{\cellcolor[HTML]{EFEFEF}\textsf{M}} & \multicolumn{1}{c|}{\cellcolor[HTML]{EFEFEF}\textsf{R}} \\
		\multicolumn{1}{|l}{\cellcolor[HTML]{C0C0C0}Human} & 97.2 & 25.0 & 73.0 & 61.7 & 91.1 & 23.2 & 68.9 & 43.1 \\
		\multicolumn{1}{|l}{\cellcolor[HTML]{C0C0C0}Majority} & 100.0 & 0.0 & 100.0 & 0.0 & 0.0 & 100.0 & 100.0 & 0.0 \\
		\multicolumn{1}{|l}{\cellcolor[HTML]{C0C0C0}SVM} & 100.0 & 26.4 & 67.8 & 62.4 & 67.3 & 77.4 & 84.9 & 55.5 \\
		\multicolumn{1}{|l}{\cellcolor[HTML]{C0C0C0}LSTM} & 98.4 & 20.8 & 57.8 & 71.5 & 61.9 & 80.2 & 86.1 & 46.8 \\
		\multicolumn{1}{|l}{\cellcolor[HTML]{C0C0C0}langBERT} & 99.2 & 41.5 & 62.0 & 73.4 & 66.0 & 59.6 & 86.7 & 58.4 \\
		\multicolumn{1}{|l}{\cellcolor[HTML]{C0C0C0}mBERT} & 100.0 & 30.3 & 61.1 & 69.1 & 66.7 & 78.8 & 81.7 & 61.9 \\
		\multicolumn{1}{|l}{\cellcolor[HTML]{C0C0C0}XLM-R} & 100.0 & 0.0 & 100.0 & 0.0 & 0.0 & 100.0 & 82.0 & 61.9 \\ \hline
	\end{tabular}
	\caption{F1 on dev for \removal, the two-way extreme speech
		classification task}
	\label{acceptable_results}
\end{table*}

\begin{table*}[]
	\centering
	\small
	\begin{tabular}{ccccc|}
		\cline{2-5}
		\rowcolor[HTML]{9B9B9B} 
		\cellcolor[HTML]{FFFFFF} & \multicolumn{1}{|c}{Brazil} & \multicolumn{1}{c}{Germany} & \multicolumn{1}{c}{India} & \multicolumn{1}{c|}{Kenya} \\
		\multicolumn{1}{|l}{\cellcolor[HTML]{C0C0C0}langBERT} & 95.4 & 92.1 & 85.5 & 83.1  \\
		\multicolumn{1}{|l}{\cellcolor[HTML]{C0C0C0}mBERT} & 94.1 & 90.3 & 92.8 & 85.6  \\
		\multicolumn{1}{|l}{\cellcolor[HTML]{C0C0C0}XLM-R} & 94.1 & 88.2 & 93.0 & 84.8  \\ \hline
	\end{tabular}
	\caption{LRAP (Label Ranking Average Precision) on
		dev for
		\target, the target group classification task}
	\label{target_classification}
\end{table*}

\begin{table*}[]
	\centering
	\small
	\begin{tabular}{lcccrrrcrrrrrr|}
		\cline{3-14}
		&\cellcolor[HTML]{FFFFFF} & \multicolumn{3}{|c}{\cellcolor[HTML]{9B9B9B}Brazil} & \multicolumn{3}{c}{\cellcolor[HTML]{9B9B9B}Germany} & \multicolumn{3}{c}{\cellcolor[HTML]{9B9B9B}India} & \multicolumn{3}{c|}{\cellcolor[HTML]{9B9B9B}Kenya} \\ 
		&\multicolumn{1}{c}{} & \multicolumn{1}{|c}{\cellcolor[HTML]{EFEFEF}Der.} & \multicolumn{1}{c}{\cellcolor[HTML]{EFEFEF}Exc.} & \multicolumn{1}{c}{\cellcolor[HTML]{EFEFEF}Dan.} & \multicolumn{1}{c}{\cellcolor[HTML]{EFEFEF}Der.} & \multicolumn{1}{c}{\cellcolor[HTML]{EFEFEF}Exc.} & \multicolumn{1}{c}{\cellcolor[HTML]{EFEFEF}Dan.} & \multicolumn{1}{c}{\cellcolor[HTML]{EFEFEF}Der.} & \multicolumn{1}{c}{\cellcolor[HTML]{EFEFEF}Exc.} & \multicolumn{1}{c}{\cellcolor[HTML]{EFEFEF}Dan.} & \multicolumn{1}{c}{\cellcolor[HTML]{EFEFEF}Der.} & \multicolumn{1}{c}{\cellcolor[HTML]{EFEFEF}Exc.} & \multicolumn{1}{c|}{\cellcolor[HTML]{EFEFEF}Dan.} \\ 
		\multirow{4}{*}{\begin{turn}{90}{\footnotesize train}\end{turn}}&		\multicolumn{1}{|l}{\cellcolor[HTML]{C0C0C0}Brazil} & 98.9 & 0.0 & 49.3 & 100.0 & 0.0 & 0.0 & 100.0 & 0.0 & 0.0 & 100.0 & 0.0 & 0.0 \\ 
		&\multicolumn{1}{|l}{\cellcolor[HTML]{C0C0C0}Germany} & 94.1 & 0.0 & 0.0 & 56.3 & 72.4 & 0.0 & 80.0 & 30.8 & 0.0 & 82.9 & 29.0 & 0.0 \\ 
		&\multicolumn{1}{|l}{\cellcolor[HTML]{C0C0C0}India} & 95.5 & 0.0 & 11.0 & 96.3 & 0.0 & 0.0 & 60.9 & 45.5 & 81.3 & 70.4 & 40.8 & 6.3 \\ 
		&\multicolumn{1}{|l}{\cellcolor[HTML]{C0C0C0}Kenya} & 94.9 & 3.0 & 9.6 & 79.6 & 10.4 & 0.0 & 83.7 & 14.4 & 29.0 & 83.5 & 48.4 & 48.8 \\ \cline{2-14}
	\end{tabular}
	\caption{F1 on dev for \extremity in cross-country transfer
		(all languages)}
	\label{cross_lingual}
\end{table*}

\begingroup
\setlength{\tabcolsep}{5pt} 
\begin{table}[]
	\centering
	\small
	\begin{tabular}{lccrrccr|}
		\cline{3-8}
		&		\multicolumn{1}{c|}{\cellcolor[HTML]{FFFFFF}} &
		\multicolumn{3}{c}{\cellcolor[HTML]{9B9B9B}IN\textsubscript{en}}
		&
		\multicolumn{3}{c|}{\cellcolor[HTML]{9B9B9B}KE\textsubscript{en}}
		\\ \cline{3-8}
		&\multicolumn{1}{c|}{} & \multicolumn{1}{c}{\cellcolor[HTML]{EFEFEF}Der.} & \multicolumn{1}{c}{\cellcolor[HTML]{EFEFEF}Exc.} & \multicolumn{1}{c}{\cellcolor[HTML]{EFEFEF}Dan.} & \multicolumn{1}{c}{\cellcolor[HTML]{EFEFEF}Der.} & \multicolumn{1}{c}{\cellcolor[HTML]{EFEFEF}Exc.} & \multicolumn{1}{c|}{\cellcolor[HTML]{EFEFEF}Dan.} \\ 
		\multirow{2}{*}{\begin{turn}{90}{\footnotesize train} \end{turn}}&		\multicolumn{1}{|l}{\cellcolor[HTML]{C0C0C0}IN\textsubscript{en}} & 60.0 & 44.8 & 0.0 & 60.9 & 50.8 & 0.0 \\ 
		&	\multicolumn{1}{|l}{\cellcolor[HTML]{C0C0C0}KE\textsubscript{en}} & 85.0 & 0.0 & 18.8 & 78.2 & 61.9 & 74.5 \\ \cline{2-8}
	\end{tabular}
	\caption{F1 on dev for \extremity for cross-country transfer in English (IN/KE = India/Kenya)}
	\label{cross_lingual_english}
\end{table}
\endgroup

\subsection{Zero-Shot Cross-Country Classification}

\subsubsection{All languages}

We evaluate \texttt{mBERT} on  zero-shot cross-country
transfer, i.e., training on one country and testing on the rest
(results are shown in Table \ref{cross_lingual}). Performance is in general poor,
indicating that \texttt{mBERT} is not able to generalize
from one country to another. Trained on Brazil, the model is
unable to make any inferences on other countries. From Kenya
to India, we see some transferability potential, with the
model correctly identifying passages in all three classes
(although at a non-competitively low rate). These results confirm our
intuition that detecting extreme speech depends on social
and cultural information, so zero-shot transfer, without
access to specific information about the target country,
is not a promising approach.

\subsubsection{English}
\label{english_english}

We investigate cross-country transfer of
\texttt{BERT}, an English model.
We only experiment with the two countries that have a
nontrivial number of English passages,
India (IN) and Kenya (KE), restricting the datasets to their English
part only (denoted by IN\textsubscript{en} and KE\textsubscript{en}, respectively). While cross-country performance is low for both countries,
we see that KE\textsubscript{en}$\rightarrow$KE\textsubscript{en} performance is high. We note that performance is better in KE\textsubscript{en}$\rightarrow$KE\textsubscript{en}
than in the previously examined
KE\textsubscript{all}$\rightarrow$KE\textsubscript{all} (where KE\textsubscript{all} is the entire Kenyan set).
This shows that for a single
language within one country, \texttt{BERT} can indeed
classify extreme speech with adequate accuracy.

\subsection{Prediction analysis with \lime}
\seclabel{interpretability}

To shed light on predictions of \texttt{mBERT} in the \extremity task (described in \secref{extremity}) we extract
top-contributing words with LIME \cite{lime}. Specifically, we
compute the words that contribute the most to \texttt{mBERT}'s predictions
(alongside their weights) for each
passage and then average the weights, returning the top 10 words
with at least 5 occurrences in the examined set. This
list is shown in \tabref{lime_explainability}.

The Indian and German sets are dominated by religious groups
(``Moslems'', ``Muslims''). In India, ethnic terms (``Rohingyas'')
are also present while in Germany we see extreme speech
targeting politicians
(``Merkel'').
In Brazil we see politically divisive terms (``Ucranizar'',
a term originally meaning ``Ukrainian Brazilian'' which has
now been appropriated to denounce opponents to the
right-wing as ``communists'') as well as insults like
``traveco'' (for ``cross-dresser'', used here as a slur). In
Kenya, we see direct insults such as ``idiot'' and
``wajinga'' (meaning ``foolish''), as well as expressions
referring to ethnic group  such as ``luo'' and ``kikuyu''.

\begin{table}[]
	\centering
	\small
	\begin{tabular}{|cccccccc|}
		\hline
		\rowcolor[HTML]{9B9B9B} 
		Brazil & Germany & \multicolumn{3}{c}{\cellcolor[HTML]{9B9B9B}India} & \multicolumn{3}{c|}{\cellcolor[HTML]{9B9B9B}Kenya} \\ 
		fechar & Politiker & \multicolumn{3}{c}{muslims} & \multicolumn{3}{c|}{cows}\\
		Ucranizar & Grünen & \multicolumn{3}{c}{Muslim} & \multicolumn{3}{c|}{ruto} \\
		ucranizar & {\scriptsize Mohammedaner} & \multicolumn{3}{c}{muslim} & \multicolumn{3}{c|}{luo} \\
		safada & Juden & \multicolumn{3}{c}{Muslims} & \multicolumn{3}{c|}{wajinga} \\
		prender & Merkels & \multicolumn{3}{c}{ko} & \multicolumn{3}{c|}{kikuyu} \\
		lixo & Merkel & \multicolumn{3}{c}{mullo} & \multicolumn{3}{c|}{stupid} \\
		coisa & Regierung & \multicolumn{3}{c}{Rohingyas} & \multicolumn{3}{c|}{idiot} \\
		kkkkk & Opfer & \multicolumn{3}{c}{\textit{Da}} & \multicolumn{3}{c|}{looting} \\
		Vagabundo & Islam & \multicolumn{3}{c}{suvar} & \multicolumn{3}{c|}{tangatanga} \\
		traveco & Moslems & \multicolumn{3}{c}{\textit{Dara}} & \multicolumn{3}{c|}{ujinga} \\ \hline
	\end{tabular}
	\caption{Top words contributing to predictions of \texttt{mBERT} for \extremity. Hindi text originally in the Devanagari script was converted to Latin characters (in italics) to comply with arxiv's typesetting policy.}
	\tablabel{lime_explainability}
\end{table}

\section{Conclusion}
\label{sec:conclusion}
We have presented
\dataset, an extreme speech dataset, containing 20,297
passages
from Brazil, Germany, India and Kenya.
We capture both granular levels of extremity and targets of
extreme speech by engaging a team of annotators from within
the affected communities. In a collaboration of
anthropologists and computational linguists, we established a community-based framework, with the goal of curating data more representative of real-world harms.

We introduce baselines for three novel tasks, including
extreme speech and target group classification. We give
experimental support for the intuition that extreme speech
classification is dependent on cultural knowledge and that
current NLP models do not capture this. Finally, we perform
interpretability analysis on \texttt{BERT}'s predictions to
reveal potential deficiencies, showing that models rely
heavily on keywords and names of marginalized groups.

We hope our community-driven work will contribute 
to the effective elimination of extreme speech 
against target groups, not just in Western democracies, but in a greater variety of countries worldwide.

\section{Acknowledgments}

This research has received funding from the European Research Council Proof of Concept grant (Agreement Number: 957442). For more about the project see \url{https://www.ai4dignity.gwi.uni-muenchen.de}.

The first and fourth authors were partly supported by the
European Research Council (\#740516).

\section{Ethical Considerations and Limitations}

\subsection{Ethics Statement}

The data provided here contains extreme speech that can be shocking and harmful. We present this dataset as a way to peel back the veil of extreme speech against the selected under-represented communities around the world. We want to motivate the analysis of this overlooked area as a whole and the investigation of the various levels of extreme speech (derogatory, exclusionary and dangerous) as found in online social media. This data is not intended and should not be used for pretraining models applied to real-world tasks, since a model pretrained on this data could potentially exhibit and propagate the extreme speech found in the passages we collected.

Further, while we endeavored to include as many communities around the world as possible, the data we collected and the list of communities we included are of course non-exhaustive. For each country, we had a close circle of annotators, therefore it is possible other marginalized groups in these countries were not covered (although we made efforts to keep this to a plausible minimum).

\eat{\subsection{Framework for Future Ethical Research}
	
	Previous work in extreme speech research as well as groundwork on data collection ethics have proposed more inclusion of marginalized communities in projects that deal with their peoples and cultures. We further advocate for this direction. Discussions with people from the researched communities were pivotal in uncovering facets of extreme speech and its targets that were previously latent.
	
	When bridging NLP with anthropology fields it is vital to draw from theories already developed and established. For example, to define extreme speech we should take inputs from fields that have worked extensively on the subject to make sure our research is in line with a broader corpus of science. The addition of anthropologists to the team brings the necessary expertise to navigate this large body of work. Without this input, we risk working in an NLP ivory tower, cutting off communications from other fields pertinent to the task at hand.}

\subsection{Limitations}

Due to limitations of both time and budget, we only gathered extreme speech without negative passages (ie. neutral language). These neutral passages form the majority of content on social media \cite{founta,social_bias_frames}. Despite the abundance of such passages, annotating them using our current scheme would be time and effort-consuming (our annotators collect data on their own, from their own networks, without us querying and supplying data to them). Thus, to keep control in the hands of annotators while at the same time keeping their workload to a reasonable minimum, we decided to only collect extreme speech passages.

\nocite{icons}
\bibliography{anthology}

\begin{thebibliography}{60}
\expandafter\ifx\csname natexlab\endcsname\relax\def\natexlab#1{#1}\fi

\bibitem[{Al~Kuwatly et~al.(2020)Al~Kuwatly, Wich, and
  Groh}]{annotator_bias_demographic}
Hala Al~Kuwatly, Maximilian Wich, and Georg Groh. 2020.
\newblock \href {https://doi.org/10.18653/v1/2020.alw-1.21} {Identifying and
  measuring annotator bias based on annotators{'} demographic characteristics}.
\newblock In \emph{Proceedings of the Fourth Workshop on Online Abuse and
  Harms}, pages 184--190, Online. Association for Computational Linguistics.

\bibitem[{Bakalis(2018)}]{cyberhate_laws}
Chara Bakalis. 2018.
\newblock \href {https://doi.org/10.1080/13600834.2017.1393934} {Rethinking
  cyberhate laws}.
\newblock \emph{Information \& Communications Technology Law}, 27(1):86--110.

\bibitem[{Banko et~al.(2020)Banko, MacKeen, and Ray}]{hatespeech_taxonomy}
Michele Banko, Brendon MacKeen, and Laurie Ray. 2020.
\newblock \href {https://doi.org/10.18653/v1/2020.alw-1.16} {A unified taxonomy
  of harmful content}.
\newblock In \emph{Proceedings of the Fourth Workshop on Online Abuse and
  Harms}, pages 125--137, Online. Association for Computational Linguistics.

\bibitem[{Bender and Friedman(2018)}]{data_statements}
Emily~M. Bender and Batya Friedman. 2018.
\newblock \href {https://doi.org/10.1162/tacl_a_00041} {Data statements for
  natural language processing: Toward mitigating system bias and enabling
  better science}.
\newblock \emph{Transactions of the Association for Computational Linguistics},
  6:587--604.

\bibitem[{Benesch(2018)}]{dangerous_speech}
Susan Benesch. 2018.
\newblock \href {https://dangerousspeech.org/guide/} {Dangerous speech: A
  practical guide}.

\bibitem[{Bleich(2014)}]{hate_speech_high_court_decisions}
Erik Bleich. 2014.
\newblock \href {https://doi.org/10.1080/1369183X.2013.851476} {Freedom of
  expression versus racist hate speech: Explaining differences between high
  court regulations in the usa and europe}.
\newblock \emph{Journal of Ethnic and Migration Studies}, 40(2):283--300.

\bibitem[{Boromisza-Habashi(2013)}]{boromisza}
David Boromisza-Habashi. 2013.
\newblock \href {https://doi.org/10.1080/10584609.2013.805682} {\emph{Speaking
  Hatefully: Culture, Communication, and Political Action in Hungary}}.
\newblock Pennsylvania State University Press.

\bibitem[{Cicchetti(1994)}]{icc}
Domenic~V. Cicchetti. 1994.
\newblock Guidelines, criteria, and rules of thumb for evaluating normed and
  standardized assessment instruments in psychology.
\newblock \emph{Psychological Assessment}, 6:284--290.

\bibitem[{Conneau et~al.(2020)Conneau, Khandelwal, Goyal, Chaudhary, Wenzek,
  Guzm{\'a}n, Grave, Ott, Zettlemoyer, and Stoyanov}]{xlmr}
Alexis Conneau, Kartikay Khandelwal, Naman Goyal, Vishrav Chaudhary, Guillaume
  Wenzek, Francisco Guzm{\'a}n, Edouard Grave, Myle Ott, Luke Zettlemoyer, and
  Veselin Stoyanov. 2020.
\newblock \href {https://doi.org/10.18653/v1/2020.acl-main.747} {Unsupervised
  cross-lingual representation learning at scale}.
\newblock In \emph{Proceedings of the 58th Annual Meeting of the Association
  for Computational Linguistics}, pages 8440--8451, Online. Association for
  Computational Linguistics.

\bibitem[{Davidson et~al.(2019{\natexlab{a}})Davidson, Bhattacharya, and
  Weber}]{racial_bias_in_data}
Thomas Davidson, Debasmita Bhattacharya, and Ingmar Weber. 2019{\natexlab{a}}.
\newblock \href {https://doi.org/10.18653/v1/W19-3504} {Racial bias in hate
  speech and abusive language detection datasets}.
\newblock In \emph{Proceedings of the Third Workshop on Abusive Language
  Online}, pages 25--35, Florence, Italy. Association for Computational
  Linguistics.

\bibitem[{Davidson et~al.(2019{\natexlab{b}})Davidson, Bhattacharya, and
  Weber}]{racial_bias_in_hatespeech}
Thomas Davidson, Debasmita Bhattacharya, and Ingmar Weber. 2019{\natexlab{b}}.
\newblock \href {https://doi.org/10.18653/v1/W19-3504} {Racial bias in hate
  speech and abusive language detection datasets}.
\newblock In \emph{Proceedings of the Third Workshop on Abusive Language
  Online}, pages 25--35, Florence, Italy. Association for Computational
  Linguistics.

\bibitem[{Davidson et~al.(2017)Davidson, Warmsley, Macy, and Weber}]{davidson}
Thomas Davidson, Dana Warmsley, Michael Macy, and Ingmar Weber. 2017.
\newblock \href {https://aaai.org/ocs/index.php/ICWSM/ICWSM17/paper/view/15665}
  {Automated hate speech detection and the problem of offensive language}.
\newblock In \emph{International AAAI Conference on Web and Social Media}.

\bibitem[{Delapouite(Accessed 10/11/2021)}]{icons}
Bruno~Heridet Delapouite. Accessed 10/11/2021.
\newblock \href {https://game-icons.net/} {https://game-icons.net/}.

\bibitem[{Donovan and danah boyd(2021)}]{stop_the_presses}
Joan Donovan and danah boyd. 2021.
\newblock \href {https://doi.org/10.1177/0002764219878229} {Stop the presses?
  moving from strategic silence to strategic amplification in a networked media
  ecosystem}.
\newblock \emph{American Behavioral Scientist}, 65(2):333--350.

\bibitem[{Founta et~al.(2018)Founta, Djouvas, Chatzakou, Leontiadis, Blackburn,
  Stringhini, Vakali, Sirivianos, and Kourtellis}]{founta}
Antigoni-Maria Founta, Constantinos Djouvas, Despoina Chatzakou, Ilias
  Leontiadis, Jeremy Blackburn, Gianluca Stringhini, Athena Vakali, Michael
  Sirivianos, and Nicolas Kourtellis. 2018.
\newblock Large scale crowdsourcing and characterization of twitter abusive
  behavior.
\newblock In \emph{11th International Conference on Web and Social Media, ICWSM
  2018}. AAAI Press.

\bibitem[{Gebru(2019)}]{gebru_gender_race}
Timnit Gebru. 2019.
\newblock \href {http://arxiv.org/abs/1908.06165} {Oxford handbook on ai ethics
  book chapter on race and gender}.

\bibitem[{Grimminger and Klinger(2021)}]{hate_towards_political_opponent}
Lara Grimminger and Roman Klinger. 2021.
\newblock \href {https://aclanthology.org/2021.wassa-1.18} {Hate towards the
  political opponent: A {T}witter corpus study of the 2020 {US} elections on
  the basis of offensive speech and stance detection}.
\newblock In \emph{Proceedings of the Eleventh Workshop on Computational
  Approaches to Subjectivity, Sentiment and Social Media Analysis}, pages
  171--180, Online. Association for Computational Linguistics.

\bibitem[{Guest et~al.(2021)Guest, Vidgen, Mittos, Sastry, Tyson, and
  Margetts}]{misogyny_annotated_data}
Ella Guest, Bertie Vidgen, Alexandros Mittos, Nishanth Sastry, Gareth Tyson,
  and Helen Margetts. 2021.
\newblock \href {https://www.aclweb.org/anthology/2021.eacl-main.114} {An
  expert annotated dataset for the detection of online misogyny}.
\newblock In \emph{Proceedings of the 16th Conference of the European Chapter
  of the Association for Computational Linguistics: Main Volume}, pages
  1336--1350, Online. Association for Computational Linguistics.

\bibitem[{Haynes(2019)}]{bolivia}
Nell Haynes. 2019.
\newblock \href {https://ijoc.org/index.php/ijoc/article/view/9109} {Writing on
  the walls: Discourses on bolivian immigrants in chilean meme humor}.
\newblock \emph{International Journal of Communication}, 13(0).

\bibitem[{Hede et~al.(2021)Hede, Agarwal, Lu, Mutz, and
  Nenkova}]{incivility_in_news}
Anushree Hede, Oshin Agarwal, Linda Lu, Diana~C. Mutz, and Ani Nenkova. 2021.
\newblock \href {https://aclanthology.org/2021.eacl-main.225} {From toxicity in
  online comments to incivility in {A}merican news: Proceed with caution}.
\newblock In \emph{Proceedings of the 16th Conference of the European Chapter
  of the Association for Computational Linguistics: Main Volume}, pages
  2620--2630, Online. Association for Computational Linguistics.

\bibitem[{Hervik(2019)}]{denmark}
Peter Hervik. 2019.
\newblock \href {https://ijoc.org/index.php/ijoc/article/view/9106} {Ritualized
  opposition in danish practices of extremist language and thought}.
\newblock \emph{International Journal of Communication}, 13(0).

\bibitem[{Jo and Gebru(2020)}]{collecting_sociocultural_data}
Eun~Seo Jo and Timnit Gebru. 2020.
\newblock \href {https://doi.org/10.1145/3351095.3372829} {Lessons from
  archives: Strategies for collecting sociocultural data in machine learning}.
\newblock In \emph{Proceedings of the 2020 Conference on Fairness,
  Accountability, and Transparency}, FAT* '20, page 306–316, New York, NY,
  USA. Association for Computing Machinery.

\bibitem[{Kim et~al.(2020)Kim, Ortiz, Nam, Santiago, and
  Datta}]{intersectional_bias}
Jae{-}Yeon Kim, Carlos Ortiz, Sarah Nam, Sarah Santiago, and Vivek Datta. 2020.
\newblock \href {http://arxiv.org/abs/2005.05921} {Intersectional bias in hate
  speech and abusive language datasets}.
\newblock \emph{CoRR}, abs/2005.05921.

\bibitem[{Krippendorff(2011)}]{alpha}
Klaus Krippendorff. 2011.
\newblock \href {https://repository.upenn.edu/asc_papers/43} {Computing
  krippendorff's alpha-reliability}.

\bibitem[{Laaksonen et~al.(2020)Laaksonen, Haapoja, Kinnunen, Nelimarkka, and
  Pöyhtäri}]{datafication_of_hate}
Salla-Maaria Laaksonen, Jesse Haapoja, Teemu Kinnunen, Matti Nelimarkka, and
  Reeta Pöyhtäri. 2020.
\newblock \href {https://doi.org/10.3389/fdata.2020.00003} {The datafication of
  hate: Expectations and challenges in automated hate speech monitoring}.
\newblock \emph{Frontiers in Big Data}, 3:3.

\bibitem[{Laugier et~al.(2021)Laugier, Pavlopoulos, Sorensen, and
  Dixon}]{civil_text_rephrasing}
L{\'e}o Laugier, John Pavlopoulos, Jeffrey Sorensen, and Lucas Dixon. 2021.
\newblock \href {https://aclanthology.org/2021.eacl-main.124} {Civil rephrases
  of toxic texts with self-supervised transformers}.
\newblock In \emph{Proceedings of the 16th Conference of the European Chapter
  of the Association for Computational Linguistics: Main Volume}, pages
  1442--1461, Online. Association for Computational Linguistics.

\bibitem[{Leins et~al.(2020)Leins, Lau, and Baldwin}]{convenience_or_death}
Kobi Leins, Jey~Han Lau, and Timothy Baldwin. 2020.
\newblock \href {https://doi.org/10.18653/v1/2020.acl-main.261} {Give me
  convenience and give her death: Who should decide what uses of {NLP} are
  appropriate, and on what basis?}
\newblock In \emph{Proceedings of the 58th Annual Meeting of the Association
  for Computational Linguistics}, pages 2908--2913, Online. Association for
  Computational Linguistics.

\bibitem[{Madukwe et~al.(2020)Madukwe, Gao, and
  Xue}]{hate_speech_data_analysis}
Kosisochukwu Madukwe, Xiaoying Gao, and Bing Xue. 2020.
\newblock \href {https://doi.org/10.18653/v1/2020.alw-1.18} {In data we trust:
  A critical analysis of hate speech detection datasets}.
\newblock In \emph{Proceedings of the Fourth Workshop on Online Abuse and
  Harms}, pages 150--161, Online. Association for Computational Linguistics.

\bibitem[{Marivate et~al.(2020)Marivate, Sefara, Chabalala, Makhaya,
  Mokgonyane, Mokoena, and Modupe}]{african_low_resource}
Vukosi Marivate, Tshephisho Sefara, Vongani Chabalala, Keamogetswe Makhaya,
  Tumisho Mokgonyane, Rethabile Mokoena, and Abiodun Modupe. 2020.
\newblock \href {https://aclanthology.org/2020.rail-1.3} {Investigating an
  approach for low resource language dataset creation, curation and
  classification: Setswana and sepedi}.
\newblock In \emph{Proceedings of the first workshop on Resources for African
  Indigenous Languages}, pages 15--20, Marseille, France. European Language
  Resources Association (ELRA).

\bibitem[{Mathew et~al.(2021)Mathew, Saha, Yimam, Biemann, Goyal, and
  Mukherjee}]{hatexplain}
Binny Mathew, Punyajoy Saha, Seid~Muhie Yimam, Chris Biemann, Pawan Goyal, and
  Animesh Mukherjee. 2021.
\newblock \href {https://ojs.aaai.org/index.php/AAAI/article/view/17745}
  {Hatexplain: A benchmark dataset for explainable hate speech detection}.
\newblock \emph{Proceedings of the AAAI Conference on Artificial Intelligence},
  35(17):14867--14875.

\bibitem[{McCoy et~al.(2019)McCoy, Pavlick, and Linzen}]{mnli_hans}
Tom McCoy, Ellie Pavlick, and Tal Linzen. 2019.
\newblock \href {https://doi.org/10.18653/v1/P19-1334} {Right for the wrong
  reasons: Diagnosing syntactic heuristics in natural language inference}.
\newblock In \emph{Proceedings of the 57th Annual Meeting of the Association
  for Computational Linguistics}, pages 3428--3448, Florence, Italy.
  Association for Computational Linguistics.

\bibitem[{McHugh(2012)}]{kappa}
Mary~L. McHugh. 2012.
\newblock \href {https://doi.org/10.11613/BM.2012.031} {Interrater reliability:
  the kappa statistic}.
\newblock \emph{Biochemia medica}, 22:276--82.

\bibitem[{Mitchell et~al.(2020)Mitchell, Baker, Moorosi, Denton, Hutchinson,
  Hanna, Gebru, and Morgenstern}]{diversity_n_inclusion_metrics}
Margaret Mitchell, Dylan Baker, Nyalleng Moorosi, Emily Denton, Ben Hutchinson,
  Alex Hanna, Timnit Gebru, and Jamie Morgenstern. 2020.
\newblock \href {https://doi.org/10.1145/3375627.3375832} {Diversity and
  inclusion metrics in subset selection}.
\newblock In \emph{Proceedings of the AAAI/ACM Conference on AI, Ethics, and
  Society}, AIES '20, page 117–123, New York, NY, USA. Association for
  Computing Machinery.

\bibitem[{Mohamed et~al.(2020)Mohamed, Png, and Isaac}]{decolonial_ai_theory}
Shakir Mohamed, Marie-Therese Png, and William Isaac. 2020.
\newblock \href {https://doi.org/10.1007/s13347-020-00405-8} {Decolonial ai:
  Decolonial theory as sociotechnical foresight in artificial intelligence}.
\newblock \emph{Philosophy and Technology}, 33(4):659–684.

\bibitem[{Nekoto et~al.(2020)Nekoto, Marivate, Matsila, Fasubaa, Fagbohungbe,
  Akinola, Muhammad, Kabongo~Kabenamualu, Osei, Sackey, Niyongabo, Macharm,
  Ogayo, Ahia, Berhe, Adeyemi, Mokgesi-Selinga, Okegbemi, Martinus, Tajudeen,
  Degila, Ogueji, Siminyu, Kreutzer, Webster, Ali, Abbott, Orife, Ezeani,
  Dangana, Kamper, Elsahar, Duru, Kioko, Espoir, van Biljon, Whitenack,
  Onyefuluchi, Emezue, Dossou, Sibanda, Bassey, Olabiyi, Ramkilowan, {\"O}ktem,
  Akinfaderin, and Bashir}]{nekoto-etal-2020-participatory}
Wilhelmina Nekoto, Vukosi Marivate, Tshinondiwa Matsila, Timi Fasubaa, Taiwo
  Fagbohungbe, Solomon~Oluwole Akinola, Shamsuddeen Muhammad, Salomon
  Kabongo~Kabenamualu, Salomey Osei, Freshia Sackey, Rubungo~Andre Niyongabo,
  Ricky Macharm, Perez Ogayo, Orevaoghene Ahia, Musie~Meressa Berhe, Mofetoluwa
  Adeyemi, Masabata Mokgesi-Selinga, Lawrence Okegbemi, Laura Martinus,
  Kolawole Tajudeen, Kevin Degila, Kelechi Ogueji, Kathleen Siminyu, Julia
  Kreutzer, Jason Webster, Jamiil~Toure Ali, Jade Abbott, Iroro Orife, Ignatius
  Ezeani, Idris~Abdulkadir Dangana, Herman Kamper, Hady Elsahar, Goodness Duru,
  Ghollah Kioko, Murhabazi Espoir, Elan van Biljon, Daniel Whitenack,
  Christopher Onyefuluchi, Chris~Chinenye Emezue, Bonaventure F.~P. Dossou,
  Blessing Sibanda, Blessing Bassey, Ayodele Olabiyi, Arshath Ramkilowan, Alp
  {\"O}ktem, Adewale Akinfaderin, and Abdallah Bashir. 2020.
\newblock \href {https://doi.org/10.18653/v1/2020.findings-emnlp.195}
  {Participatory research for low-resourced machine translation: A case study
  in {A}frican languages}.
\newblock In \emph{Findings of the Association for Computational Linguistics:
  EMNLP 2020}, pages 2144--2160, Online. Association for Computational
  Linguistics.

\bibitem[{Nozza(2021)}]{hatespeech_zero_shot_cross_lingual}
Debora Nozza. 2021.
\newblock \href {https://doi.org/10.18653/v1/2021.acl-short.114} {Exposing the
  limits of zero-shot cross-lingual hate speech detection}.
\newblock In \emph{Proceedings of the 59th Annual Meeting of the Association
  for Computational Linguistics and the 11th International Joint Conference on
  Natural Language Processing (Volume 2: Short Papers)}, pages 907--914,
  Online. Association for Computational Linguistics.

\bibitem[{Ousidhoum et~al.(2019)Ousidhoum, Lin, Zhang, Song, and
  Yeung}]{multi_hate_speech}
Nedjma Ousidhoum, Zizheng Lin, Hongming Zhang, Yangqiu Song, and Dit-Yan Yeung.
  2019.
\newblock \href {https://doi.org/10.18653/v1/D19-1474} {Multilingual and
  multi-aspect hate speech analysis}.
\newblock In \emph{Proceedings of the 2019 Conference on Empirical Methods in
  Natural Language Processing and the 9th International Joint Conference on
  Natural Language Processing (EMNLP-IJCNLP)}, pages 4675--4684, Hong Kong,
  China. Association for Computational Linguistics.

\bibitem[{Posch et~al.(2018)Posch, Bleier, Flöck, and
  Strohmaier}]{global_workforce_demographics}
Lisa Posch, Arnim Bleier, Fabian Flöck, and Markus Strohmaier. 2018.
\newblock Characterizing the global crowd workforce: A cross-country comparison
  of crowdworker demographics.

\bibitem[{Price et~al.(2020)Price, Gifford-Moore, Flemming, Musker, Roichman,
  Sylvain, Thain, Dixon, and Sorensen}]{unhealthy_convo_attributes}
Ilan Price, Jordan Gifford-Moore, Jory Flemming, Saul Musker, Maayan Roichman,
  Guillaume Sylvain, Nithum Thain, Lucas Dixon, and Jeffrey Sorensen. 2020.
\newblock \href {https://doi.org/10.18653/v1/2020.alw-1.15} {Six attributes of
  unhealthy conversations}.
\newblock In \emph{Proceedings of the Fourth Workshop on Online Abuse and
  Harms}, pages 114--124, Online. Association for Computational Linguistics.

\bibitem[{Ranasinghe and Zampieri(2020)}]{hatespeech_crosslingual_embds}
Tharindu Ranasinghe and Marcos Zampieri. 2020.
\newblock \href {https://doi.org/10.18653/v1/2020.emnlp-main.470} {Multilingual
  offensive language identification with cross-lingual embeddings}.
\newblock In \emph{Proceedings of the 2020 Conference on Empirical Methods in
  Natural Language Processing (EMNLP)}, pages 5838--5844, Online. Association
  for Computational Linguistics.

\bibitem[{Ribeiro et~al.(2016)Ribeiro, Singh, and Guestrin}]{lime}
Marco~Tulio Ribeiro, Sameer Singh, and Carlos Guestrin. 2016.
\newblock \href {https://doi.org/10.1145/2939672.2939778} {"why should i trust
  you?": Explaining the predictions of any classifier}.
\newblock In \emph{Proceedings of the 22nd ACM SIGKDD International Conference
  on Knowledge Discovery and Data Mining}, KDD '16, page 1135–1144, New York,
  NY, USA. Association for Computing Machinery.

\bibitem[{Risch and Krestel(2020)}]{risch-krestel-2020-bagging}
Julian Risch and Ralf Krestel. 2020.
\newblock \href {https://aclanthology.org/2020.trac-1.9} {Bagging {BERT} models
  for robust aggression identification}.
\newblock In \emph{Proceedings of the Second Workshop on Trolling, Aggression
  and Cyberbullying}, pages 55--61, Marseille, France. European Language
  Resources Association (ELRA).

\bibitem[{Rogers(2020)}]{deplatforming}
Richard Rogers. 2020.
\newblock \href {https://doi.org/10.1177/0267323120922066} {Deplatforming:
  Following extreme internet celebrities to telegram and alternative social
  media}.
\newblock \emph{European Journal of Communication}, 35(3):213--229.

\bibitem[{Ross et~al.(2017)Ross, Rist, Carbonell, Cabrera, Kurowsky, and
  Wojatzki}]{refugee_crisis_german}
Bj{\"{o}}rn Ross, Michael Rist, Guillermo Carbonell, Benjamin Cabrera, Nils
  Kurowsky, and Michael Wojatzki. 2017.
\newblock \href {http://arxiv.org/abs/1701.08118} {Measuring the reliability of
  hate speech annotations: The case of the european refugee crisis}.
\newblock \emph{CoRR}, abs/1701.08118.

\bibitem[{R{\"o}ttger et~al.(2021)R{\"o}ttger, Vidgen, Nguyen, Waseem,
  Margetts, and Pierrehumbert}]{hatecheck}
Paul R{\"o}ttger, Bertie Vidgen, Dong Nguyen, Zeerak Waseem, Helen Margetts,
  and Janet Pierrehumbert. 2021.
\newblock \href {https://doi.org/10.18653/v1/2021.acl-long.4} {{H}ate{C}heck:
  Functional tests for hate speech detection models}.
\newblock In \emph{Proceedings of the 59th Annual Meeting of the Association
  for Computational Linguistics and the 11th International Joint Conference on
  Natural Language Processing (Volume 1: Long Papers)}, pages 41--58, Online.
  Association for Computational Linguistics.

\bibitem[{Saltman and Russell(2014)}]{white_paper}
Erin~Marie Saltman and Jonathan Russell. 2014.
\newblock \href
  {https://preventviolentextremism.info/white-paper-%E2%80%93-role-prevent-countering-online-extremism}
  {White paper – the role of prevent in countering online extremism}.

\bibitem[{Sap et~al.(2019)Sap, Card, Gabriel, Choi, and
  Smith}]{aae_bias_hatespeech}
Maarten Sap, Dallas Card, Saadia Gabriel, Yejin Choi, and Noah~A. Smith. 2019.
\newblock \href {https://doi.org/10.18653/v1/P19-1163} {The risk of racial bias
  in hate speech detection}.
\newblock In \emph{Proceedings of the 57th Annual Meeting of the Association
  for Computational Linguistics}, pages 1668--1678, Florence, Italy.
  Association for Computational Linguistics.

\bibitem[{Sap et~al.(2020)Sap, Gabriel, Qin, Jurafsky, Smith, and
  Choi}]{social_bias_frames}
Maarten Sap, Saadia Gabriel, Lianhui Qin, Dan Jurafsky, Noah~A. Smith, and
  Yejin Choi. 2020.
\newblock \href {https://doi.org/10.18653/v1/2020.acl-main.486} {Social bias
  frames: Reasoning about social and power implications of language}.
\newblock In \emph{Proceedings of the 58th Annual Meeting of the Association
  for Computational Linguistics}, pages 5477--5490, Online. Association for
  Computational Linguistics.

\bibitem[{Shmueli et~al.(2021)Shmueli, Fell, Ray, and
  Ku}]{ethical_implications_crowdsourcing}
Boaz Shmueli, Jan Fell, Soumya Ray, and Lun-Wei Ku. 2021.
\newblock \href {https://doi.org/10.18653/v1/2021.naacl-main.295} {Beyond fair
  pay: Ethical implications of {NLP} crowdsourcing}.
\newblock In \emph{Proceedings of the 2021 Conference of the North American
  Chapter of the Association for Computational Linguistics: Human Language
  Technologies}, pages 3758--3769, Online. Association for Computational
  Linguistics.

\bibitem[{Swamy et~al.(2019)Swamy, Jamatia, and
  Gamb{\"a}ck}]{hate_speech_cross_dataset}
Steve~Durairaj Swamy, Anupam Jamatia, and Bj{\"o}rn Gamb{\"a}ck. 2019.
\newblock \href {https://doi.org/10.18653/v1/K19-1088} {Studying
  generalisability across abusive language detection datasets}.
\newblock In \emph{Proceedings of the 23rd Conference on Computational Natural
  Language Learning (CoNLL)}, pages 940--950, Hong Kong, China. Association for
  Computational Linguistics.

\bibitem[{Udupa(2021)}]{counter_online_hate}
Sahana Udupa. 2021.
\newblock \href
  {http://nbn-resolving.de/urn/resolver.pl?urn=nbn:de:bvb:19-epub-77473-7}
  {Digital technology and extreme speech: Approaches to counter online hate}.

\bibitem[{Udupa et~al.(2021)Udupa, Hickok, Maronikolakis, Schuetze, Csuka,
  Wisiorek, and Nann}]{policy_brief}
Sahana Udupa, Elonnai Hickok, Antonis Maronikolakis, Hinrich Schuetze, Laura
  Csuka, Axel Wisiorek, and Leah Nann. 2021.
\newblock \href
  {http://nbn-resolving.de/urn/resolver.pl?urn=nbn:de:bvb:19-epub-76087-9}
  {Artificial intelligence, extreme speech and the challenges of online content
  moderation}.

\bibitem[{Udupa and Pohjonen(2019)}]{extreme_speech_sahana}
Sahana Udupa and Matti Pohjonen. 2019.
\newblock \href {https://ijoc.org/index.php/ijoc/article/view/9102} {Extreme
  speech and global digital cultures}.
\newblock \emph{International Journal of Communication}, 13(0).

\bibitem[{Vidgen et~al.(2019)Vidgen, Harris, Nguyen, Tromble, Hale, and
  Margetts}]{hate_speech_frontiers}
Bertie Vidgen, Alex Harris, Dong Nguyen, Rebekah Tromble, Scott Hale, and Helen
  Margetts. 2019.
\newblock \href {https://doi.org/10.18653/v1/W19-3509} {Challenges and
  frontiers in abusive content detection}.
\newblock In \emph{Proceedings of the Third Workshop on Abusive Language
  Online}, pages 80--93, Florence, Italy. Association for Computational
  Linguistics.

\bibitem[{Waseem(2016)}]{are_you_racist}
Zeerak Waseem. 2016.
\newblock \href {https://doi.org/10.18653/v1/W16-5618} {Are you a racist or am
  {I} seeing things? annotator influence on hate speech detection on
  {T}witter}.
\newblock In \emph{Proceedings of the First Workshop on {NLP} and Computational
  Social Science}, pages 138--142, Austin, Texas. Association for Computational
  Linguistics.

\bibitem[{Waseem et~al.(2017)Waseem, Davidson, Warmsley, and
  Weber}]{understanding_abuse}
Zeerak Waseem, Thomas Davidson, Dana Warmsley, and Ingmar Weber. 2017.
\newblock \href {https://doi.org/10.18653/v1/W17-3012} {Understanding abuse: A
  typology of abusive language detection subtasks}.
\newblock In \emph{Proceedings of the First Workshop on Abusive Language
  Online}, pages 78--84, Vancouver, BC, Canada. Association for Computational
  Linguistics.

\bibitem[{Waseem and Hovy(2016)}]{waseem}
Zeerak Waseem and Dirk Hovy. 2016.
\newblock \href {https://doi.org/10.18653/v1/N16-2013} {Hateful symbols or
  hateful people? predictive features for hate speech detection on {T}witter}.
\newblock In \emph{Proceedings of the {NAACL} Student Research Workshop}, pages
  88--93, San Diego, California. Association for Computational Linguistics.

\bibitem[{Wiegand et~al.(2021)Wiegand, Ruppenhofer, and
  Eder}]{implicitly_abusive_language}
Michael Wiegand, Josef Ruppenhofer, and Elisabeth Eder. 2021.
\newblock \href {https://www.aclweb.org/anthology/2021.naacl-main.48}
  {Implicitly abusive language {--} what does it actually look like and why are
  we not getting there?}
\newblock In \emph{Proceedings of the 2021 Conference of the North American
  Chapter of the Association for Computational Linguistics: Human Language
  Technologies}, pages 576--587, Online. Association for Computational
  Linguistics.

\bibitem[{Wiegand et~al.(2019)Wiegand, Ruppenhofer, and
  Kleinbauer}]{hatespeech_biased_datasets}
Michael Wiegand, Josef Ruppenhofer, and Thomas Kleinbauer. 2019.
\newblock \href {https://doi.org/10.18653/v1/N19-1060} {{D}etection of
  {A}busive {L}anguage: the {P}roblem of {B}iased {D}atasets}.
\newblock In \emph{Proceedings of the 2019 Conference of the North {A}merican
  Chapter of the Association for Computational Linguistics: Human Language
  Technologies, Volume 1 (Long and Short Papers)}, pages 602--608, Minneapolis,
  Minnesota. Association for Computational Linguistics.

\bibitem[{Zoph et~al.(2016)Zoph, Yuret, May, and
  Knight}]{zoph-etal-2016-transfer}
Barret Zoph, Deniz Yuret, Jonathan May, and Kevin Knight. 2016.
\newblock \href {https://doi.org/10.18653/v1/D16-1163} {Transfer learning for
  low-resource neural machine translation}.
\newblock In \emph{Proceedings of the 2016 Conference on Empirical Methods in
  Natural Language Processing}, pages 1568--1575, Austin, Texas. Association
  for Computational Linguistics.

\end{thebibliography}
\bibliographystyle{acl_natbib}

\clearpage
\appendix

\section{Data Statement}
\label{app:data_statement}

\hspace{\parindent}{\fontfamily{qpl}\selectfont\textit{CURATION RATIONAL}} ~ In our project, we venture to present a dataset on extreme speech across different countries (Brazil, Germany, India and Kenya).
Fact-checkers from these countries were requested to gather and annotate data. These fact-checkers searched online platforms and communities to identify extreme speech based on their contextual language.
The choice of sources was left to the fact-checkers, since they have intimate knowledge of the spread of extreme speech. Sources include social media (e.g., Twitter), fora (e.g., groups on Telegram) and direct messaging.

{\fontfamily{qpl}\selectfont\textit{LANGUAGE VARIETY}} ~ Data was collected for Brazilian Portuguese (pt-BR), German (de-DE), Hindi (hi-IN, either in the Devangari or Latin script), Swahili (sw-KE) and English used as a second language alongside these native languages.

{\fontfamily{qpl}\selectfont\textit{SPEAKER DEMOGRAPHICS}} ~ Speaker demographics were not recorded (and anonymized where necessary). Data was collected from Brazil, Germany, India and Kenya, so a fair assumption is that speakers come from these countries.

{\fontfamily{qpl}\selectfont\textit{ANNOTATOR DEMOGRAPHICS}} ~ Annotators were accredited fact-checkers in their respective countries. There were 8 female and 5 male annotators (per country, female/male counts are 2/1 in Brazil, 4/0 in Germany, 2/2 in India and 0/2 in Kenya). They were native speakers of (Brazilian) Portuguese, German, Hindi and Swahili. Ages were not recorded. Further (self-disclosed) information on annotators can be found at \url{https://www.ai4dignity.gwi.uni-muenchen.de/partnering-fact-checkers/}.

\eat{In terms of ethnicities, there were 3 Brazilians, 4 Germans, 4 Indians (embedded in communities spanning the North, Central and South India) and 2 Kenyans (both Luo, one from traditional Luo territory and the other from the capital, Nairobi).}

{\fontfamily{qpl}\selectfont\textit{SPEECH SITUATION}} ~ Speech consists entirely of text, posted and collected in 2020 and 2021. Text is mainly asynchronous, informal and spontaneous. Certain passages were posted as responses to other text (which was not collected) in a more synchronous manner. By the nature of this project, all passages contain a level of extremity.

{\fontfamily{qpl}\selectfont\textit{TEXT CHARACTERISTICS}} ~ Text comes from social media in the form of user comments. Length was limited to approximately two paragraphs (at the discretion of the annotators).

{\fontfamily{qpl}\selectfont\textit{OTHER}} ~ The team spanned multiple disciplines, ages and ethnicities.

\section{Data Analysis}

\subsection{Institutions of Power}

Statistics of institutions of power are shown in Table \ref{power_dist}. These groups can only be the target of derogatory speech, since we want to avoid censoring of speech aimed at these groups. Across all countries, we see that politicians are the predominant targets.

\subsection{Average Passage Length}

In Table \ref{avg_len} we show the average length of passages per label for each country. All sets show similar lengths, except Brazil where passages are overall shorter. Also, across sets, the more extreme a passage is, the longer it is on average.

\begin{table}[]
	\centering
	\small
	\begin{tabular}{lccccc|}
		\cline{2-6}
		& \cellcolor[HTML]{9B9B9B}Brazil & \cellcolor[HTML]{9B9B9B}Germany & \cellcolor[HTML]{9B9B9B}India & \cellcolor[HTML]{9B9B9B}Kenya & \cellcolor[HTML]{9B9B9B}\textit{Total} \\
		\multicolumn{1}{|l}{\cellcolor[HTML]{C0C0C0}Der.} & 15.8 & 22.5 & 26.0 & 24.2 & \textit{21.0} \\
		\multicolumn{1}{|l}{\cellcolor[HTML]{C0C0C0}Exc.} & 18.3 & 27.7 & 28.1 & 27.6 & \textit{27.6} \\
		\multicolumn{1}{|l}{\cellcolor[HTML]{C0C0C0}Dan.} & 21.2 & 40.5 & 30.3 & 29.6 & \textit{29.3} \\
		\multicolumn{1}{|l}{\cellcolor[HTML]{C0C0C0}Ovr.} & 16.1 & 25.0 & 27.8 & 25.7 & \textit{23.5} \\ \hline
	\end{tabular}
	\caption{Average passage length statistics}
	\label{avg_len}
\end{table}

\section{Annotation Details}
\label{app:annotation}

\subsection{Logistics}
\label{app:logistics}

There are at least two annotators from each country. In some countries, we worked with fact-checker teams which themselves employ multiple fact-checkers. In these instances, annotation work was split according to the requirements and resources of the particular team. We ensured that all involved members were accredited fact-checkers and were interviewed by our anthropology team to verify they are familiar with extreme speech and are capable of identifying it. Payment was 1.5 Euros per passage provided for the original dataset and 1 Euro per passage for the re-annotation task.

\subsection{Cross-annotation}
\label{app:cross_annotation}

In Table \ref{inter_annotator_dist} we show the number of passages cross-annotated by each annotator. Annotators were split into two groups, A and B, according to availability and were tasked with cross-annotating the passages provided by the other group.

\subsection{Inter-annotator agreement details}

In Table \ref{inter_annotator_table_all} we show inter-annotator agreement scores per country. While Germany and Kenya have acceptable scores, the other two countries have low scores.

\begin{table}[]
	\centering
	\small
	\begin{tabular}{lccclclclcl|}
		\cline{2-3}
		& \multicolumn{4}{|c}{\cellcolor[HTML]{9B9B9B}Group A} & \multicolumn{6}{c|}{\cellcolor[HTML]{9B9B9B}Group B} \\
		\multicolumn{1}{|l}{\cellcolor[HTML]{C0C0C0}Brazil} & \multicolumn{2}{c}{834} & \multicolumn{2}{c}{833} & \multicolumn{6}{c|}{833} \\
		\multicolumn{1}{|l}{\cellcolor[HTML]{C0C0C0}Germany} & \multicolumn{2}{c}{834} & \multicolumn{2}{c}{833} & \multicolumn{6}{c|}{833} \\
		\multicolumn{1}{|l}{\cellcolor[HTML]{C0C0C0}India} & \multicolumn{4}{c}{1250} & \multicolumn{6}{c|}{417 417 416} \\
		\multicolumn{1}{|l}{\cellcolor[HTML]{C0C0C0}Kenya} & \multicolumn{4}{c}{1250} & \multicolumn{6}{c|}{1250} \\ \hline
	\end{tabular}
	\caption{Number of passages each group of annotators cross-annotated, evenly split among the members of each group. Details in Appendix \ref{app:cross_annotation}.}
	\label{inter_annotator_dist}
\end{table}

\subsection{Online Interface}
\label{app:online_interface}

\begin{figure*}
	\centering
	\includegraphics[scale=0.8]{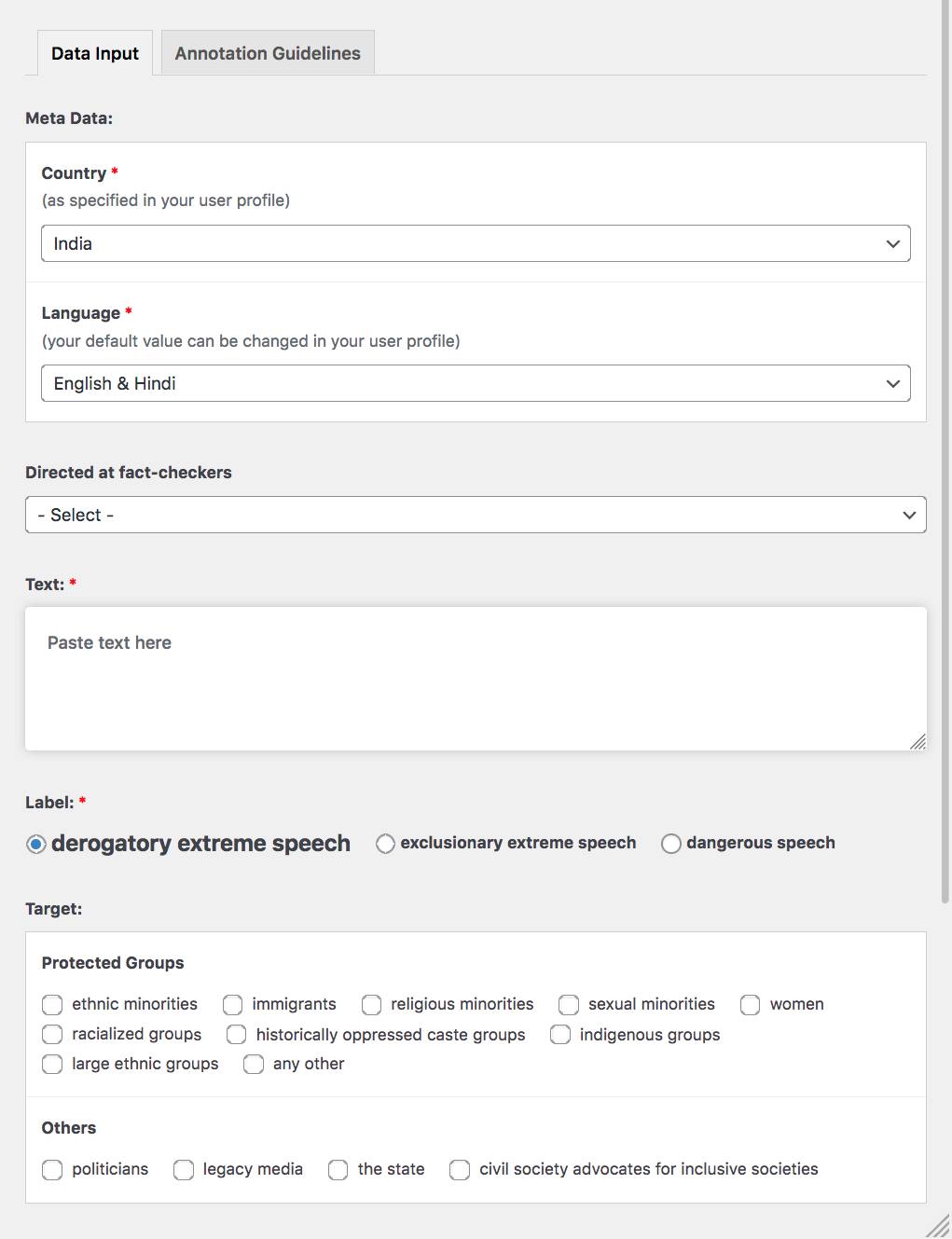}
	\caption{Interface presented to the annotators for data entry and labeling}
	\label{interface}
\end{figure*}

In Figure \ref{interface} we see the interface annotators used to enter and annotate data.

\section{Reannotation}
\label{app:reannotation}

After discussion with the annotators from Kenya, we found that there was disagreement surrounding two ethnic groups and the power dynamics around them. Namely, the Kikuyu and Kalenjin, two ethnic groups currently in power in Kenya. They make up around 17\% (largest group) and 13\% (third largest group) of the population of Kenya, respectively. Because of their position of power, in a lot of sociopolitical issues these two ethnic groups (either jointly or individually) get pitted against the rest of the population. So, in that binary perspective (e.g., Kikuyus vs.\ ``others''), the ethnic group in power was considered an ethnic minority by one annotator. The other annotator did not share this perspective and labeled these ethnic groups as indigenous groups. After a series of discussions with the annotators, a consensus was reached that the ethnic groups in power will be labeled neither as ethnic minorities nor as indigenous groups, but as a new target label: ``large ethnic groups''. This entailed that re-labeling of the extremity of these passages should take place.

\section{Model Details}
\label{app:models}

Transformer models were finetuned for 3 epochs (5 minutes each), LSTMs for 5 and SVMs until convergence. A maximum length of 128 was used universally. For each baseline, three runs were made with results averaged. Standard deviations were minimal and were not reported for brevity.

The \texttt{BERT}-based models we used are:\footnote{\url{https://huggingface.co/models}}

\begin{enumerate}
	\item \texttt{bert-base-multilingual-cased}: \url{https://huggingface.co/bert-base-multilingual-cased}
	\item \texttt{bert-base-portuguese-cased}: \url{https://huggingface.co/neuralmind/bert-base-portuguese-cased}
	\item \texttt{bert-base-german-cased}: \url{https://huggingface.co/bert-base-german-cased}
	\item \texttt{hindi-bert}: \url{https://huggingface.co/monsoon-nlp/hindi-bert}
	\item \texttt{bert-base-uncased-swahili}: \url{https://huggingface.co/flax-community/bert-base-uncased-swahili}
\end{enumerate}

\section{Combined Multilingual Setting}
\label{app:multilingual}

We perform an ablation study by combining all sets across countries and repeating our \texttt{mBERT} experiments in this new multilingual task (Table \ref{multilingual_setting_res}).

\begin{table}[]
	\centering
	\small
	\begin{tabular}{lccc|}
		\cline{2-4}
		\multicolumn{1}{c|}{\cellcolor[HTML]{FFFFFF}} & \multicolumn{3}{c}{\cellcolor[HTML]{9B9B9B}All} \\ 
		& \multicolumn{1}{c}{\cellcolor[HTML]{EFEFEF}Der.} & \multicolumn{1}{c}{\cellcolor[HTML]{EFEFEF}Exc.} & \cellcolor[HTML]{EFEFEF}Dan. \\
		\multicolumn{1}{l}{\cellcolor[HTML]{C0C0C0}mBERT} & \multicolumn{1}{c}{84.9} & \multicolumn{1}{c}{55.1} & \multicolumn{1}{c|}{50.4} \\
		& \multicolumn{1}{c}{\cellcolor[HTML]{EFEFEF}\textsf{M}} & \multicolumn{2}{c|}{\cellcolor[HTML]{EFEFEF}\textsf{R}} \\
		\multicolumn{1}{l}{\cellcolor[HTML]{C0C0C0}mBERT} & \multicolumn{1}{c}{85.5} & \multicolumn{2}{c|}{56.8} \\
		& \multicolumn{3}{c|}{\cellcolor[HTML]{EFEFEF}Target Group} \\
		\multicolumn{1}{l}{\cellcolor[HTML]{C0C0C0}mBERT} & \multicolumn{3}{c|}{91.4} \\ \cline{2-4}
	\end{tabular}
	\caption{Combined multilingual setting results.}
	\label{multilingual_setting_res}
\end{table}

Even though the use of a ``catch-all'' model that is able to work on all languages sounds enticing, care should be taken to ensure that the model has sufficient understanding for each language and culture instead of making predictions based on dubious statistical cues \cite{mnli_hans}. This is a task out of scope for this work, but we are adding such a model to our baselines for completion.

\section{Test Set Results}
\label{app:test_set}

In Tables \ref{test_set_extreme}, \ref{test_set_acceptable}, \ref{test_target_classification}, \ref{test_cross_lingual} and \ref{test_cross_lingual_english} we show results on the test set for tasks defined in \secref{experiments}.

\begin{table*}[]
	\centering
	\small
	\begin{tabular}{r|rrccrrrrrr|}
		\cline{2-11}
		\rowcolor[HTML]{9B9B9B} 
		\cellcolor[HTML]{FFFFFF} & \multicolumn{1}{|c}{\cellcolor[HTML]{9B9B9B}$\kappa$} & \multicolumn{1}{c}{$\alpha$} & \multicolumn{1}{c}{$\mbox{ICC}(3,k)$} & \multicolumn{1}{c}{Targets} & \multicolumn{1}{c}{Ovr.} & \multicolumn{1}{c}{Der.} & \multicolumn{1}{c}{\cellcolor[HTML]{9B9B9B}Exc.} & \multicolumn{1}{c}{Dan.} & \multicolumn{1}{c}{\textsf{M}} & \multicolumn{1}{c|}{\textsf{R}}  \\
		\multicolumn{1}{|l}{\cellcolor[HTML]{C0C0C0}\textit{Overall}} & \textit{0.23} & \textit{0.24} & \textit{0.41} & \textit{0.69} & \textit{63.0} & \textit{78.4} & \textit{40.2} & \textit{18.8} & \textit{78.4} & \textit{46.3} \\
		\multicolumn{1}{|l}{\cellcolor[HTML]{C0C0C0}Brazil} & 0.08 & 0.12 & 0.19 & 0.62 & 85.9 & 91.3 & 12.7 & 5.8 & 91.3 & 6.7 \\
		\multicolumn{1}{|l}{\cellcolor[HTML]{C0C0C0}Germany} & 0.35 & 0.35 & 0.52 & 0.79 & 68.2 & 73.0 & 61.6 & 0.0  & 73.0 & 61.7 \\
		\multicolumn{1}{|l}{\cellcolor[HTML]{C0C0C0}India} & 0.11 & 0.04 & 0.19 & 0.81 & 39.6 & 72.2 & 30.2 & 5.3  & 72.2 & 39.7 \\
		\multicolumn{1}{|l}{\cellcolor[HTML]{C0C0C0}Kenya} & 0.13 & 0.21 & 0.47 & 0.50 & 58.1 & 69.4 & 11.8 & 57.1 & 69.4 & 43.0 \\ \hline
	\end{tabular}
	\caption{Inter-annotator agreement table. In order, $\kappa$, $\alpha$ and $\mbox{ICC}(3, k)$ for extreme speech labels, target groups ($\kappa$), overall accuracy (\%), derogatory/exclusionary/dangerous (\%), \textsf{M}/\textsf{R} (\%)}
	\label{inter_annotator_table_all}
\end{table*}

\begin{table*}[]
	\centering
	\small
	\begin{tabular}{lcccccccccc}
		\cline{2-6}
		& \multicolumn{2}{|c}{\cellcolor[HTML]{9B9B9B}Brazil} & \multicolumn{2}{c}{\cellcolor[HTML]{9B9B9B}Germany} & \multicolumn{2}{c}{\cellcolor[HTML]{9B9B9B}India} & \multicolumn{2}{c}{\cellcolor[HTML]{9B9B9B}Kenya} & \multicolumn{2}{c}{\cellcolor[HTML]{9B9B9B}\textit{Total}} \\ 
		\multicolumn{1}{l}{\cellcolor[HTML]{FFFFFF}} & \multicolumn{1}{|c}{\cellcolor[HTML]{EFEFEF}n} & \multicolumn{1}{c}{\cellcolor[HTML]{EFEFEF}\%} & \multicolumn{1}{c}{\cellcolor[HTML]{EFEFEF}n} & \multicolumn{1}{c}{\cellcolor[HTML]{EFEFEF}\%} & \multicolumn{1}{c}{\cellcolor[HTML]{EFEFEF}n} & \multicolumn{1}{c}{\cellcolor[HTML]{EFEFEF}\%} & \multicolumn{1}{c}{\cellcolor[HTML]{EFEFEF}n} & \multicolumn{1}{c}{\cellcolor[HTML]{EFEFEF}\%} & \multicolumn{1}{c}{\cellcolor[HTML]{EFEFEF}\textit{n}} & \multicolumn{1}{c|}{\cellcolor[HTML]{EFEFEF}\textit{\%}} \\
		\multicolumn{1}{|l}{\cellcolor[HTML]{C0C0C0}Politicians} & \multicolumn{1}{r}{1105} & \multicolumn{1}{r}{59.6} & \multicolumn{1}{r}{778} & \multicolumn{1}{r}{69.8} & \multicolumn{1}{r}{273} & \multicolumn{1}{r}{67.6} & \multicolumn{1}{r}{2098} & \multicolumn{1}{r}{93.9} & \multicolumn{1}{r}{\textit{4254}} & \multicolumn{1}{r|}{\textit{75.9}} \\
		\multicolumn{1}{|l}{\cellcolor[HTML]{C0C0C0}Legacy Media} & \multicolumn{1}{r}{663} & \multicolumn{1}{r}{35.8} & \multicolumn{1}{r}{106} & \multicolumn{1}{r}{9.5} & \multicolumn{1}{r}{75} & \multicolumn{1}{r}{18.6} & \multicolumn{1}{r}{54} & \multicolumn{1}{r}{2.4} & \multicolumn{1}{r}{\textit{898}} & \multicolumn{1}{r|}{\textit{16.0}} \\
		\multicolumn{1}{|l}{\cellcolor[HTML]{C0C0C0}The State} & \multicolumn{1}{r}{55} & \multicolumn{1}{r}{3.0} & \multicolumn{1}{r}{171} & \multicolumn{1}{r}{15.4} & \multicolumn{1}{r}{20} & \multicolumn{1}{r}{5.0} & \multicolumn{1}{r}{74} & \multicolumn{1}{r}{3.3} & \multicolumn{1}{r}{\textit{320}} & \multicolumn{1}{r|}{\textit{5.7}} \\
		\multicolumn{1}{|l}{\cellcolor[HTML]{C0C0C0}Civil Society Advocates} & \multicolumn{1}{r}{30} & \multicolumn{1}{r}{1.6} & \multicolumn{1}{r}{59} & \multicolumn{1}{r}{5.3} & \multicolumn{1}{r}{36} & \multicolumn{1}{r}{8.9} & \multicolumn{1}{r}{9} & \multicolumn{1}{r}{0.4} & \multicolumn{1}{r}{\textit{134}} & \multicolumn{1}{r|}{\textit{2.4}} \\ \hline
	\end{tabular}
	\caption{Distribution of institutions of power as targets of derogatory extreme speech, in total numbers (n) and percentages (\%)}
	\label{power_dist}
\end{table*}

\begin{table*}[]
	\centering
	\small
	\begin{tabular}{crrrrrcrrrrrr|}
		\cline{2-13}
		\rowcolor[HTML]{9B9B9B} 
		\cellcolor[HTML]{FFFFFF} & \multicolumn{3}{|c}{\cellcolor[HTML]{9B9B9B}Brazil} & \multicolumn{3}{c}{\cellcolor[HTML]{9B9B9B}Germany} & \multicolumn{3}{c}{\cellcolor[HTML]{9B9B9B}India} & \multicolumn{3}{c|}{\cellcolor[HTML]{9B9B9B}Kenya} \\ 
		\rowcolor[HTML]{EFEFEF} 
		\multicolumn{1}{l|}{\cellcolor[HTML]{FFFFFF}} & \multicolumn{1}{c}{\cellcolor[HTML]{EFEFEF}Der.} & \multicolumn{1}{c}{\cellcolor[HTML]{EFEFEF}Exc.} & \multicolumn{1}{c}{\cellcolor[HTML]{EFEFEF}Dan.} & \multicolumn{1}{c}{\cellcolor[HTML]{EFEFEF}Der.} & \multicolumn{1}{c}{\cellcolor[HTML]{EFEFEF}Exc.} & \multicolumn{1}{c}{\cellcolor[HTML]{EFEFEF}Dan.} & \multicolumn{1}{c}{\cellcolor[HTML]{EFEFEF}Der.} & \multicolumn{1}{c}{\cellcolor[HTML]{EFEFEF}Exc.} & \multicolumn{1}{c}{\cellcolor[HTML]{EFEFEF}Dan.} & \multicolumn{1}{c}{\cellcolor[HTML]{EFEFEF}Der.} & \multicolumn{1}{c}{\cellcolor[HTML]{EFEFEF}Exc.} & \multicolumn{1}{c|}{\cellcolor[HTML]{EFEFEF}Dan.} \\
		\multicolumn{1}{|l}{\cellcolor[HTML]{C0C0C0}SVM} & \multicolumn{1}{r}{99.7} & \multicolumn{1}{r}{2.7} & \multicolumn{1}{r}{27.7} & \multicolumn{1}{r}{68.7} & \multicolumn{1}{r}{65.8} & \multicolumn{1}{r}{0.0} & \multicolumn{1}{r}{66.8} & \multicolumn{1}{r}{34.6} & \multicolumn{1}{r}{70.3} & \multicolumn{1}{r}{91.4} & \multicolumn{1}{r}{35.6} & \multicolumn{1}{r|}{34.3}\\
		\multicolumn{1}{|l}{\cellcolor[HTML]{C0C0C0}LSTM} & \multicolumn{1}{r}{98.7} & \multicolumn{1}{r}{0.8} & \multicolumn{1}{r}{0.0} & \multicolumn{1}{r}{78.2} & \multicolumn{1}{r}{55.9} & \multicolumn{1}{r}{0.0} & \multicolumn{1}{r}{54.5} & \multicolumn{1}{r}{62.6} & \multicolumn{1}{r}{0.0} & \multicolumn{1}{r}{66.8} & \multicolumn{1}{r}{68.2} & \multicolumn{1}{r|}{0.0} \\
		\multicolumn{1}{|l}{\cellcolor[HTML]{C0C0C0}langBERT} & \multicolumn{1}{r}{99.7} & \multicolumn{1}{r}{2.7} & \multicolumn{1}{r}{37.7} & \multicolumn{1}{r}{71.1} & \multicolumn{1}{r}{69.5} & \multicolumn{1}{r}{0.0} & \multicolumn{1}{r}{85.6} & \multicolumn{1}{r}{6.6} & \multicolumn{1}{r}{74.4} & \multicolumn{1}{r}{83.3} & \multicolumn{1}{r}{38.5} & 45.3 \\
		\multicolumn{1}{|l}{\cellcolor[HTML]{C0C0C0}mBERT} & \multicolumn{1}{r}{99.5} & \multicolumn{1}{r}{0.0} & \multicolumn{1}{r}{34.8} & \multicolumn{1}{r}{58.2} & \multicolumn{1}{r}{74.0} & \multicolumn{1}{r}{0.0} & \multicolumn{1}{r}{93.1} & \multicolumn{1}{r}{4.1} & \multicolumn{1}{r}{73.6} & \multicolumn{1}{r}{86.2} & \multicolumn{1}{r}{47.1} & 55.2 \\
		\multicolumn{1}{|l}{\cellcolor[HTML]{C0C0C0}XLM-R} & \multicolumn{1}{r}{100.0} & \multicolumn{1}{r}{0.0} & \multicolumn{1}{r}{0.0} & \multicolumn{1}{r}{65.6} & \multicolumn{1}{r}{76.2} & \multicolumn{1}{c}{0.0} & \multicolumn{1}{r}{96.3} & \multicolumn{1}{r}{0.0} & \multicolumn{1}{r}{49.6} & \multicolumn{1}{r}{90.6} & \multicolumn{1}{r}{35.3} & 24.4 \\ \hline
	\end{tabular}
	\caption{F1 for \extremity, the three-way extreme speech
		classification task on the test set}
	\label{test_set_extreme}
\end{table*}

\begin{table*}[]
	\centering
	\small
	\begin{tabular}{
			>{\columncolor[HTML]{C0C0C0}}crrrrrrrr|}
		\cline{2-9}
		\cellcolor[HTML]{FFFFFF} & \multicolumn{2}{|c}{\cellcolor[HTML]{9B9B9B}Brazil} & \multicolumn{2}{c}{\cellcolor[HTML]{9B9B9B}Germany} & \multicolumn{2}{c}{\cellcolor[HTML]{9B9B9B}India} & \multicolumn{2}{c|}{\cellcolor[HTML]{9B9B9B}Kenya} \\ 
		\multicolumn{1}{l|}{\cellcolor[HTML]{FFFFFF}} & \multicolumn{1}{c}{\cellcolor[HTML]{EFEFEF}\textsf{M}} & \multicolumn{1}{c}{\cellcolor[HTML]{EFEFEF}\textsf{R}} & \multicolumn{1}{c}{\cellcolor[HTML]{EFEFEF}\textsf{M}} & \multicolumn{1}{c}{\cellcolor[HTML]{EFEFEF}\textsf{R}} & \multicolumn{1}{c}{\cellcolor[HTML]{EFEFEF}\textsf{M}} & \multicolumn{1}{c}{\cellcolor[HTML]{EFEFEF}\textsf{R}} & \multicolumn{1}{c}{\cellcolor[HTML]{EFEFEF}\textsf{M}} & \multicolumn{1}{c|}{\cellcolor[HTML]{EFEFEF}\textsf{R}} \\
		\multicolumn{1}{|l}{\cellcolor[HTML]{C0C0C0}SVM} & \multicolumn{1}{r}{99.7} & 19.3 & \multicolumn{1}{r}{68.3} & 67.4 & \multicolumn{1}{r}{57.8} & 76.3 & \multicolumn{1}{r}{87.3} & 53.8  \\
		\multicolumn{1}{|l}{\cellcolor[HTML]{C0C0C0}LSTM} & \multicolumn{1}{r}{97.6} & 24.8 & \multicolumn{1}{r}{78.6} & 52.0 & \multicolumn{1}{r}{64.7} & 80.3 & \multicolumn{1}{r}{82.4} & 56.7 \\
		\multicolumn{1}{|l}{\cellcolor[HTML]{C0C0C0}langBERT} & \multicolumn{1}{r}{99.7} & 29.3 & \multicolumn{1}{r}{72.3} & 69.3 & \multicolumn{1}{r}{71.9} & 76.1 & \multicolumn{1}{r}{86.7} & 50.8 \\
		\multicolumn{1}{|l}{\cellcolor[HTML]{C0C0C0}mBERT} & \multicolumn{1}{r}{100.0} & 0.0 & \multicolumn{1}{r}{54.2} & 75.9 & \multicolumn{1}{r}{80.0} & 50.6 & \multicolumn{1}{r}{86.5} & 61.4 \\
		\multicolumn{1}{|l}{\cellcolor[HTML]{C0C0C0}XLM-R} & \multicolumn{1}{r}{100.0} & 0.0 & \multicolumn{1}{r}{100.0} & 0.0 & \multicolumn{1}{r}{0.0} & 100.0 & \multicolumn{1}{r}{86.5} & 63.2 \\ \hline
	\end{tabular}
	\caption{F1 for \removal, the two-way extreme speech
		classification task on the test set}
	\label{test_set_acceptable}
\end{table*}

\begin{table*}[]
	\centering
	\small
	\begin{tabular}{ccccc|}
		\cline{2-5}
		\rowcolor[HTML]{9B9B9B} 
		\cellcolor[HTML]{FFFFFF} & \multicolumn{1}{|c}{Brazil} & \multicolumn{1}{c}{Germany} & \multicolumn{1}{c}{India} & \multicolumn{1}{c|}{Kenya} \\
		\multicolumn{1}{|l}{\cellcolor[HTML]{C0C0C0}langBERT} & 95.7 & 91.0 & 82.3 & 86.0  \\
		\multicolumn{1}{|l}{\cellcolor[HTML]{C0C0C0}mBERT} & 95.2 & 90.0 & 91.7 & 89.3  \\
		\multicolumn{1}{|l}{\cellcolor[HTML]{C0C0C0}XLM-R} & 95.2  & 89.9 & 90.1 & 87.2  \\ \hline
	\end{tabular}
	\caption{LRAP (Label Ranking Average Precision) for
		\target, the target group classification task on the test set}
	\label{test_target_classification}
\end{table*}

\begin{table*}[]
	\centering
	\small
	\begin{tabular}{lcccrrrcrrrrrr|}
		\cline{3-14}
		&\cellcolor[HTML]{FFFFFF} & \multicolumn{3}{|c}{\cellcolor[HTML]{9B9B9B}Brazil} & \multicolumn{3}{c}{\cellcolor[HTML]{9B9B9B}Germany} & \multicolumn{3}{c}{\cellcolor[HTML]{9B9B9B}India} & \multicolumn{3}{c}{\cellcolor[HTML]{9B9B9B}Kenya} \\ 
		&\multicolumn{1}{c}{} & \multicolumn{1}{|c}{\cellcolor[HTML]{EFEFEF}Der.} & \multicolumn{1}{c}{\cellcolor[HTML]{EFEFEF}Exc.} & \multicolumn{1}{c}{\cellcolor[HTML]{EFEFEF}Dan.} & \multicolumn{1}{c}{\cellcolor[HTML]{EFEFEF}Der.} & \multicolumn{1}{c}{\cellcolor[HTML]{EFEFEF}Exc.} & \multicolumn{1}{c}{\cellcolor[HTML]{EFEFEF}Dan.} & \multicolumn{1}{c}{\cellcolor[HTML]{EFEFEF}Der.} & \multicolumn{1}{c}{\cellcolor[HTML]{EFEFEF}Exc.} & \multicolumn{1}{c}{\cellcolor[HTML]{EFEFEF}Dan.} & \multicolumn{1}{c}{\cellcolor[HTML]{EFEFEF}Der.} & \multicolumn{1}{c}{\cellcolor[HTML]{EFEFEF}Exc.} & \multicolumn{1}{c|}{\cellcolor[HTML]{EFEFEF}Dan.} \\ 
		\multirow{4}{*}{\begin{turn}{90}{\footnotesize train}\end{turn}}&		\multicolumn{1}{|l}{\cellcolor[HTML]{C0C0C0}Brazil} & 99.5 & 0.0 & 34.8 & 100.0 & 0.0 & 0.0 & 100.0 & 0.0 & 0.0 & 100.0 & 0.0 & 0.0 \\ 
		&\multicolumn{1}{|l}{\cellcolor[HTML]{C0C0C0}Germany} & 82.6 & 18.9 & 0.0 & 58.2 & 74.0 & 0.0 & 62.5 & 49.2 & 0.0 & 82.1 & 22.1 & 0.0 \\ 
		&\multicolumn{1}{|l}{\cellcolor[HTML]{C0C0C0}India} & 63.9 & 5.4 & 31.9 & 56.2 & 37.2 & 0.0 & 93.1 & 4.1 & 73.6 & 69.7 & 34.6 & 9.0 \\ 
		&\multicolumn{1}{|l}{\cellcolor[HTML]{C0C0C0}Kenya} & 95.2 & 0.0 & 2.9 & 82.7 & 7.2 & 0.0 & 79.4 & 8.2 & 32.0 & 90.6 & 35.3 & 24.4 \\ \cline{2-14}
	\end{tabular}
	\caption{F1 for \extremity in cross-country transfer
		(all languages) on the test set}
	\label{test_cross_lingual}
\end{table*}

\begin{table*}[]
	\centering
	\small
	\begin{tabular}{lccrrccr|}
		\cline{3-8}
		&		\cellcolor[HTML]{FFFFFF} &
		\multicolumn{3}{|c}{\cellcolor[HTML]{9B9B9B}IN$_{en}$}
		&
		\multicolumn{3}{c}{\cellcolor[HTML]{9B9B9B}KE$_{en}$}
		\\ \cline{3-8}
		&\multicolumn{1}{c}{} & \multicolumn{1}{|c}{\cellcolor[HTML]{EFEFEF}Der.} & \multicolumn{1}{c}{\cellcolor[HTML]{EFEFEF}Exc.} & \multicolumn{1}{c}{\cellcolor[HTML]{EFEFEF}Dan.} & \multicolumn{1}{c}{\cellcolor[HTML]{EFEFEF}Der.} & \multicolumn{1}{c}{\cellcolor[HTML]{EFEFEF}Exc.} & \multicolumn{1}{c|}{\cellcolor[HTML]{EFEFEF}Dan.} \\ 
		\multirow{2}{*}{\begin{turn}{90}{\footnotesize train} \end{turn}}&		\multicolumn{1}{|l}{\cellcolor[HTML]{C0C0C0}IN\textsubscript{en}} & 60.0 & 69.0 & 50.0 & 62.1 & 45.4 & 0.0 \\ 
		&	\multicolumn{1}{|l}{\cellcolor[HTML]{C0C0C0}KE\textsubscript{en}} & 83.3 & 4.0 & 18.8 & 84.3 & 62.1 & 55.1 \\ \cline{2-8}
	\end{tabular}
	\caption{F1 for \extremity for cross-country transfer
		in English on the test set (IN/KE = India/Kenya)}
	\label{test_cross_lingual_english}
\end{table*}

\end{document}